\documentclass[journal]{IEEEtran}
\usepackage{blindtext}
\usepackage{graphicx}

\usepackage{graphicx}
\usepackage{color}
\usepackage{caption}
\usepackage{subcaption}
\usepackage{subfloat}
\usepackage{multirow}
\usepackage{amsmath}
\usepackage{bm}
\usepackage{enumerate}
\usepackage{pdflscape}
\usepackage{booktabs,caption,fixltx2e}
\usepackage[flushleft]{threeparttable}
\usepackage{placeins}
\usepackage{multirow}

\date{}
\usepackage{hyperref}
\usepackage{tikz}
\usetikzlibrary{positioning,shapes}

\ifCLASSINFOpdf
\else
\fi
\begin{document}
\title{Applications of Trajectory Data from the Perspective of a Road Transportation Agency: Literature Review and Maryland Case Study}

\author{Nikola~Markovi\'c,~Przemys\l{}aw~Seku\l{}a,~Zachary~Vander~Laan,~Gennady~Andrienko,~and~Natalia~Andrienko% <-this % stops a space
\thanks{The first three authors are with the Center for Advanced Transportation Technology, Department of Civil and Environmental Engineering, University of Maryland, College Park, MD, USA, while the last two authors are with the Fraunhofer Institute for Intelligent Analysis and Information Systems, Sankt Augustin, Germany and the City University London, UK. The second author is also affiliated with the University of Economics in Katowice, Poland.}% <-this % stops a space
\thanks{Manuscript received December 7, 2017, revised April 6, 2018; }}%revised January 11, 2007.}}


\markboth{IEEE TRANSACTIONS ON INTELLIGENT TRANSPORTATION SYSTEMS}%
{Shell \MakeLowercase{\textit{et al.}}: Bare Demo of IEEEtran.cls for Journals}



\maketitle

\begin{abstract}
Transportation agencies have an opportunity to leverage increasingly-available trajectory datasets to  improve their analyses and decision-making processes.  However, this data is typically purchased from vendors, which means agencies must understand its potential benefits beforehand in order to properly assess its value relative to the cost of acquisition. While the literature concerned with trajectory data is rich, it is naturally fragmented and focused on technical contributions in niche areas, which makes it difficult for government agencies to assess its value across different transportation domains. To overcome this issue, the current paper explores trajectory data from the perspective of a road transportation agency interested in acquiring trajectories to enhance its analyses. The paper provides a literature review illustrating applications of trajectory data in six areas of road transportation systems analysis: demand estimation, modeling human behavior, designing public transit, traffic performance measurement and prediction, environment and safety. In addition, it visually explores 20 million GPS traces in Maryland, illustrating existing and suggesting new applications of trajectory data.
\end{abstract}


% \fi
\IEEEpeerreviewmaketitle

\section{Introduction}
Numerous detailed trajectory datasets have recently become available, including Global Positioning System (GPS) traces from cell phones and vehicles, anonymized Call Detail Records (CDR) from cell phone providers, and data from arrays of Bluetooth and Wi-Fi detectors that re-identify devices over time. As vast amounts of spatiotemporal data becomes more ubiquitous, transportation agencies have an opportunity to leverage these resources to improve analysis techniques and answer important questions more efficiently. However, since this data often needs to be purchased, agencies should be well-informed about potential benefits in order to assess its value to their organization. While the literature concerned with trajectory data is rich, it is naturally fragmented and often focuses on technical contributions in niche areas, which makes it hard for government agencies to assess its specific application to transportation domains. To overcome this issue, the current paper explores trajectory data from the perspective of a transportation agency, seeking to synthesize existing approaches and also present new applications for transportation systems analysis. It is worth noting that a recent review paper \cite{andrienko2017visual} also seeks to bring trajectory data closer to practice; however, it focuses in particular on \textit{visual analytics} approaches that may be useful for transportation agencies. The authors conclude that it is necessary to establish collaboration between the visual analytics and transportation research communities, and seek to do so in the current paper.

The trajectories analyzed in this paper were obtained from a major GPS company in North America. It provides Internet services and mobile applications informing users about traffic conditions, which are estimated based on terabytes of GPS data collected daily from millions of mobile phones, cars, trucks and other fleet vehicles. In 2016 the Maryland State Highway Administration (SHA) purchased GPS traces of all trips recorded in Maryland during four months of the previous year. The SHA subsequently asked the authors of this paper to determine the value of the trajectory data for transportation systems analysis and evaluate the cost/benefit trade-off. With the goal of enabling SHA and other government agencies to accurately assess the value of these datasets, this paper provides an overview of potential use-cases in various domains of transportation engineering. In particular, we make two contributions:

\begin{itemize}
	\item We provide a literature review illustrating innovative uses of trajectory data in road transportation systems analysis. The review includes studies that exploit different trajectory datasets (GPS traces, CDR, Bluetooth and Wi-Fi detectors) in six areas of transportation engineering: demand estimation, modeling human behavior, designing public transit, traffic performance measurement and prediction, environmental impact, and safety analysis. This review can serve as a single reference point for government agencies trying to decide whether purchasing trajectory data would be beneficial to their multifaceted analyses.
	
	\item We visually explore a set of 20 million GPS traces in Maryland, demonstrating existing and suggesting new applications of trajectory data in road transportation systems analysis. The suggested novel applications include: (a) design of isochrones via density-based clustering/filtering of trajectory data that can be applied without any information about the underlying transportation network and historical travel times along different road links, and (b) weight/speed enforcement that could improve safety and reduce property damage while employing simple processing and visualization of trajectory data. Lastly, we summarize the best-practices in analyzing trajectories and discuss data-related challenges that transportation agencies should be aware of when purchasing data. 
\end{itemize}

The next section provides a literature review illustrating various uses of trajectory data in road transportation, while the following section showcases its application in Maryland. After discussing data-related challenges, we conclude by summarizing the findings. 

\section{Literature Review}\label{sec:LitReview}
Applications of trajectory data in road transportation are synthesized into six areas, each of which is discussed in a separate subsection. It is worth noting that the following review does not seek to provide an exhaustive overview of the literature, but highlight some of the relevant work in order to illustrate applications of trajectory data in different areas of transportation engineering. The review generally focuses on relatively recent papers that include comprehensive case studies, which could be of particular interest to transportation agencies.

\subsection{Demand estimation}\label{sec:demandlit}
At the core of demand modeling and transportation planning is the problem of estimating the number of trips that take place between specific locations \cite{Iqbal2014}. The traditional data sources used to estimate demand are census and travel survey data, which are sometimes combined with traffic counts from roadside sensors. While these datasets contain valuable information, their use is sometimes hindered by non-representative sampling and misreported responses on surveys, as well as difficulties with reconstructing the trips between Origin-Destination (O-D) pairs based on sparse vehicle count data. Given the importance of determining O-D pairs and the shortcomings of traditional methods, mobility data offers an appealing, more direct approach to inferring demand. 

\subsubsection{An example O-D matrix derivation based on trajectories}
The methodology employed in \cite{Toole2015} is an innovative example of how trajectory data can be utilized to estimate demand. The proposed approach begins with a preprocessing procedure, where CDR data points representing timestamps of phone calls and text messages are mapped to locations, either through triangulation methods or simply by locating the nearest cell tower. Individual anonymous users' locations are then tracked over time to form trajectories through space, making the data functionally similar to GPS trajectory data, but with less spatial resolution. From this point, the goal is to mine the data to extract the number of trips that take place between locations, a process that involves making assumptions about how to define important locations and assign meaning to the set of movements over time. The authors use an algorithm from \cite{zheng2011learning} to transform the detailed trajectory data into more manageable trajectories of stay locations, where a stay location represents a place in which a cell-phone user spends significant amounts of time. The region is then divided into a set of zones, and stay points are assigned to the zone that encompasses them, meaning that the stay-point trajectories represent trips between zones. After carefully discarding users who do not use their phones frequently enough to accurately characterize their travel behavior and scaling the results to reflect the total population, an estimate of daily O-D trip tables can be produced. 

\subsubsection{Additional considerations}
It should be noted that other demand models consider time-of-day effects (e.g., AM/PM Peak, Off Peak) and the types of trips taken (e.g., Home-Work, Home-Other), which can also be extracted from mobility data. These considerations will be discussed in the next subsection.

\subsection{Modeling human behavior}
Quantifying human behavior is a key component of demand modeling and transportation planning, since understanding why people travel and the specific choices they make in the process (e.g., mode and route choice) can be useful for shaping policies that positively impact the overall transportation system. This subsection focuses on two specific aspects of human behavior: assigning context to travel movements and choice analysis.

\subsubsection{Context of travel movement}
Given detailed mobility datasets, intelligent data mining strategies can be utilized to derive meaning and context from the locations visited. A recent paper \cite{Zheng2015} provides a thorough overview of the field, distilling trajectory data mining into the following phases: (a) preprocessing (trajectory compression, stay-point detection, trajectory segmentation and map matching), (b) data management (indexing and storing data so it can be retrieved quickly) and (c) pattern mining (clustering by time/shape/segment, classifying, and detecting outliers). The last phase is particularly interesting for transportation, because its application involves grouping similar trip origins, destinations, times of day, trip durations, and sections of road, in order to extract prevailing patterns and answer transportation-related questions. 

The aspect of trajectory mining most relevant to demand estimation is the stay-point detection process, which helps identify locations at which an individual spends significant amounts of time. For example, \cite{andrienko2016scalable} uses trajectory data to detect and classify significant locations (e.g., home, work, or social) in a way that respects user privacy, employing a visual analytics approach and demonstrating its capabilities on a benchmark dataset and location data from Twitter. Another example includes \cite{Schneider2013}, which utilizes the concept of network motifs to investigate and describe the types of locations where cell phone users spend extended periods of time.

\subsubsection{Choice analysis}
The other aspect of modeling human behavior that can be improved through mobility data is describing choice behavior (e.g., mode and route choice). In transportation, people's behavior is usually addressed with discrete choice models, where users consider a set of mutually exclusive alternatives and choose the one that maximizes their utility. Given a set of observations about travel behavior from some segment of the population, a transportation modeler seeks to find model parameters that best describe the observed behavior \cite{Ben-Akiva1985}. Consequently, detailed travel survey data is vital to discrete choice modeling, but is laborious to acquire and may become outdated after a few years. Accordingly, mobility data provides an opportunity to observe how people behave, from which discrete choice models can be calibrated, verified, or shown to be flawed. This can be illustrated through the following two studies.

A recent work \cite{Xu2016} combines CDR, Waze GPS data and a handful of other sources to investigate the impact of special events on a city's travel patterns, focusing on the 2016 Olympics in Rio de Janeiro, Brazil. The authors estimate O-D demand prior to the Olympics, and creatively utilize the Olympic event schedule, stadium capacities, Airbnb and hotel information to account for additional destinations and demand from tourists. After building the demand model to account for the Olympics, they explore choice behavior in the form of mode shift and traffic routing strategies, noting the overall system implications associated with the different choice behaviors. Another example is \cite{Lima2016}, where the authors use GPS traces of 526 vehicles to investigate routing behavior and check whether people take the lowest-cost paths, which is commonly assumed in traffic assignment. They cluster origins and destinations to find important locations, cluster trajectories to determine possible routes, and discover that most users take the same path in the majority of situations, which often is not the minimum cost path. Studies like these help determine whether choice models that are based on utility maximization actually match real-world behavior. 

\subsection{Designing public transit}\label{sec:transitlit}
Public transit systems provide an effective way to help relieve congestion, reduce emissions, and transport people efficiently in areas where significant travel demand exists between common origins or destinations \cite{Pinelli2016}. Transit planning consists of selecting system characteristics (e.g., station locations, routes, fleet size, service frequencies, fares) in order to provide satisfactory service at minimal cost \cite{Guihaire2008}. This task can be aided by trajectory data in different ways.

\subsubsection{Trajectories as input to optimization models}
The traditional transportation network optimization techniques rely on aggregate O-D matrices \cite{Pinelli2016}, which we have already discussed in the demand estimation subsection of this paper. We reemphasize that, in addition to traditional survey/land use/traffic count methods, these O-D matrices can be estimated by mining trajectory data. Note that this approach may be particularly useful in developing countries where survey data may not be available, and in cities where travel survey data quickly becomes outdated due to rapid population growth. In such cases, trajectory data may help provide reasonable aggregate demand estimates to feed existing transit network optimization models. An example of this approach is found in \cite{berlingerio2013allaboard}, where the authors use CDR to propose route changes to a transit system in Abidjan, Ivory Coast, resulting in estimated average travel time reductions of up to 10\% across the city.

\subsubsection{Data-driven approach}
There is another, more data-driven approach to transit planning. Rather than reducing trajectory data to a set of important O-D locations and using these O-D matrices to feed an optimization model, the data-driven approach seeks to use the trajectory data directly to infer optimal transit routes. One of the most complete examples of this approach is found in \cite{Pinelli2016}, where the authors propose a methodology to design a new transit network in Abidjan, Ivory Coast, using the aforementioned cell phone data. The premise is based on the idea that a transit network's service should reflect the spatial and temporal patterns of people's movement. Based on patterns that emerge from the massive amount of cell phone data points (referred to as m-trails), a set of potential routes are selected and then refined by employing other utility-maximization strategies. Upon selecting the routes, the authors use linear programming to find optimal service frequencies.

\subsection{Traffic performance measurement and prediction}
Trajectory data can be used both to analyze historical performance of a traffic system and to help predict future traffic states. Upon discussing the related work, we point out existing challenges in this area.

\subsubsection{Quantifying past performance}
Transportation agencies require traffic data in order to quantify system performance, inform policy decisions, and identify areas of improvement \cite{BrennanJr.2013}. Important performance indicators include congestion-related measures, such as travel times over different time periods, travel time reliability, vehicle/person throughput, occupancy, and total vehicle delay, all of which depend on the ability to accurately capture data. Traditional traffic sensors such as induction loop detectors and radar/microwave detectors are useful for obtaining vehicle counts, but have more difficulty estimating travel time distributions because these fixed sensors measure only spot-mean speed \cite{kesting2013traffic}. There are many intelligent techniques that can be used to overcome this drawback of traditional detector data, but trajectory datasets offer an alternative, direct approach for measuring travel times. Rather than inferring travel times based on point measurements and constant-speed assumptions, these datasets can be used directly to calculate travel time distributions, quantify congestion measures, and serve as a ground truth for other sensor data \cite{BrennanJr.2016}. State and local agencies can leverage these probe vehicle data and existing methodologies to develop mobility reports. 

\subsubsection{Real-time predictions}
In addition to quantifying past performance of a transportation network, traffic data can be used for real-time traffic state predictions, provided that data feeds are available in real time. With some exceptions (e.g., \cite{Wedin2015}), literature in this area tends to focus on data assimilation techniques, which seek to optimally blend predictions from traffic models and field measurement observations, each of which contain some unknown levels of uncertainty \cite{evensen2009data}. While significant data assimilation research has been performed using stationary sensors, trajectory-based measurements provide new opportunities for traffic state estimation. For example, \cite{Wei2010} investigates the performance of a Kalman filtering approach to travel time estimation using data from a fleet of GPS-enabled probe vehicles, with traditional traffic sensors serving as ground truth measurements. Recognizing that GPS and loop detector data sources contain complementary information, others consider assimilation techniques that merge data collected from both fixed and moving measurements, while focusing on different aspects of the assimilation problem and application areas (e.g., \cite{Yang2005} considers arterial traffic in the context of disruptive events). 

\subsection{Environment}
The transportation sector was responsible for 26\% of all 2014 greenhouse gas emissions in the United States, mostly from burning fossil fuel for vehicles, trains, planes, and ships \cite{epa2016inventory}. Thus, transportation agencies are often interested in (a) quantifying their environmental impact, and (b) developing strategies to make operations more efficient and shift reliance away from fossil fuels. Both can be aided by the use of trajectory data.

\subsubsection{Quantifying emissions}
An important way to quantify the environmental impact of traffic is through transportation emissions models, which can be approached from macroscopic or microscopic vantage points. Macro-level models (e.g., EMEP, EEA) base the emissions calculations on aggregate flows and average vehicle speeds along transportation networks \cite{Bandeira2014}. In contrast, micro-level models focus on individual vehicles' accelerations and decelerations, which produce more accurate emissions estimates than macroscopic models, an example of which includes VT-Micro \cite{Rakha2004}. Since trajectory data can be used to improve demand estimation techniques, its application to macroscopic emissions modeling yields more accurate estimates. Similarly, since micro-level emissions models rely on knowledge of vehicle accelerations, trajectory data can be used to calculate these inputs directly rather than relying on estimates from microsimulation experiments (e.g., \cite{Feng2011}). From either perspective, trajectory data provides an opportunity to better quantify existing emissions resulting from transportation operations. 

\subsubsection{Mitigating emissions}
One attempt to reduce green house gas emissions is by developing vehicles which use alternative energy sources. Electric vehicles are one such alternative, but are hindered by a lack of necessary infrastructure for conveniently recharging. Thus, in an attempt to promote adoption of electric vehicle and related technologies, cities and planning agencies may be interested in determining how to best locate recharging/refueling infrastructure. A handful of recent studies suggest that trajectory data may be beneficial for achieving these goals, including \cite{yang2017data}. This work uses taxi GPS traces from China as input to a facility location model, seeking to determine optimal locations and capacities of charging facilities. Another attempt to reduce emissions is to use trajectory data to enable efficient carpooling \cite{berlingerio2017graal}, which can be done by extracting mobility patterns from data and using those in an optimization setting to minimize the number of cars needed for carpooling.

\subsection{Safety}
Trajectory data has recently been used in a number of innovative applications focusing on emergency response and cyclist/pedestrian safety. 

\subsubsection{Emergency response}
A recent paper \cite{Hara2015} demonstrates how trajectory data may be useful during emergencies by using probe vehicle and smartphone GPS data to assess network conditions after a 2011 earthquake in Japan and makes recommendations for disaster management. In response to the devastating earthquake, \cite{song2014intelligent} develops a methodology for probabilistically modeling human movement using GPS traces to help better respond to future disasters. Similarly, \cite{ikeda2016evacuation} determines optimal evacuation routes after natural disasters, employing a multi-objective genetic algorithm to jointly optimize evacuation distance, time, and safety.

\subsubsection{Cyclist and pedestrian safety}
A separate branch of safety research leverages GPS, Wi-Fi and Bluetooth trajectory data to provide insight into cyclist and pedestrian safety. For example, \cite{dozza2014introducing} investigates bicycle risk by analyzing GPS traces, calculating incident rates through simple odds ratios, and concluding that crash risk is greatest at intersections and on roads that are in poor condition. A related research combines GPS traces with bicycle count data to infer high-risk areas for cycling injuries \cite{Strauss2015}. These analyses provide methodological frameworks and recommendations that may be useful for transportation agencies looking to design bike lanes or improve bikeshare safety.

From a pedestrian and urban planning perspective, \cite{koshak2008analyzing} uses GPS traces to characterize human movement in order to address the issue of excessive pedestrian density during special religious events in Saudi Arabia. Likewise, \cite{johansson2010analysis} analyzes trajectories during a crowd disaster to characterize how pedestrian dynamics change from low to unsafe crowd densities. Although the empirical data is extracted from video, it is nonetheless trajectory data that can be treated similarly to datasets collected from other technologies.

\section{Maryland Case Study}
In this section we showcase several applications of trajectory data in road transportation. Most importantly, we propose three innovative applications that (to the best of our knowledge) have not been considered in the literature: measuring accessibility via density-based clustering/filtering of waypoints, identifying candidate locations for speed cameras, and selecting regions for additional vehicle weight enforcement. In addition, we illustrate applications of trajectory data in estimating demand and evaluating transit systems that have been extensively addressed in the literature (see Sections \ref{sec:demandlit} and \ref{sec:transitlit}). These are included because they are highly applicable to many transportation agencies, and help illustrate how the results can be effectively communicated to practitioners. Also, we note that analysis of demand is relevant for other applications discussed in the literature review, which highlights the importance of  inferring overall traffic volumes from raw trajectory data.

\subsection{Data}\label{sec:Data}
\begin{figure}
	\centering
	\begin{subfigure}[b]{0.16\textwidth}
		\centering		
		\frame{\includegraphics[height=24mm]{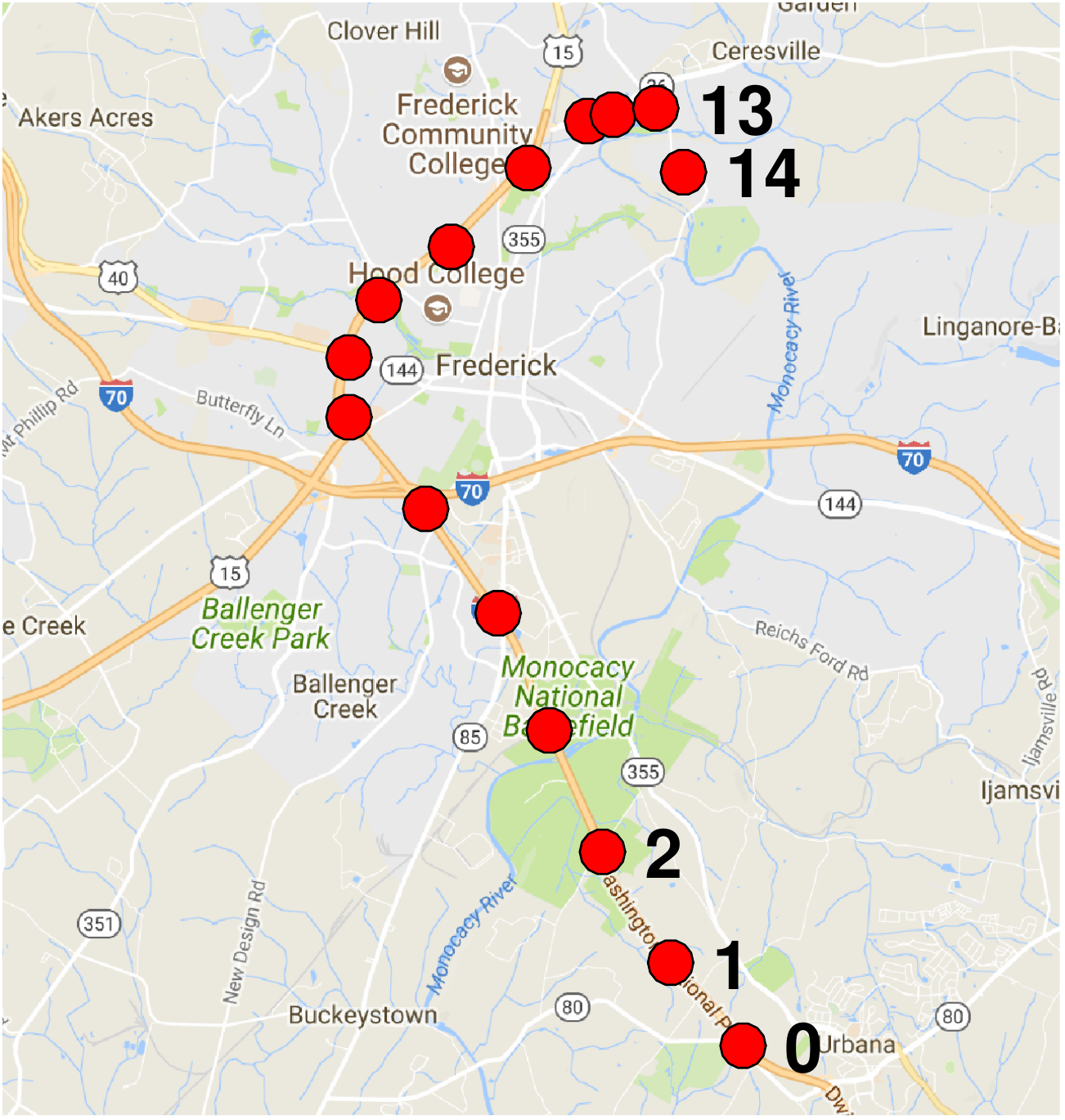}}
	\end{subfigure}%
	\begin{subfigure}[b]{0.16\textwidth}
		\centering
		\includegraphics[height=24mm]{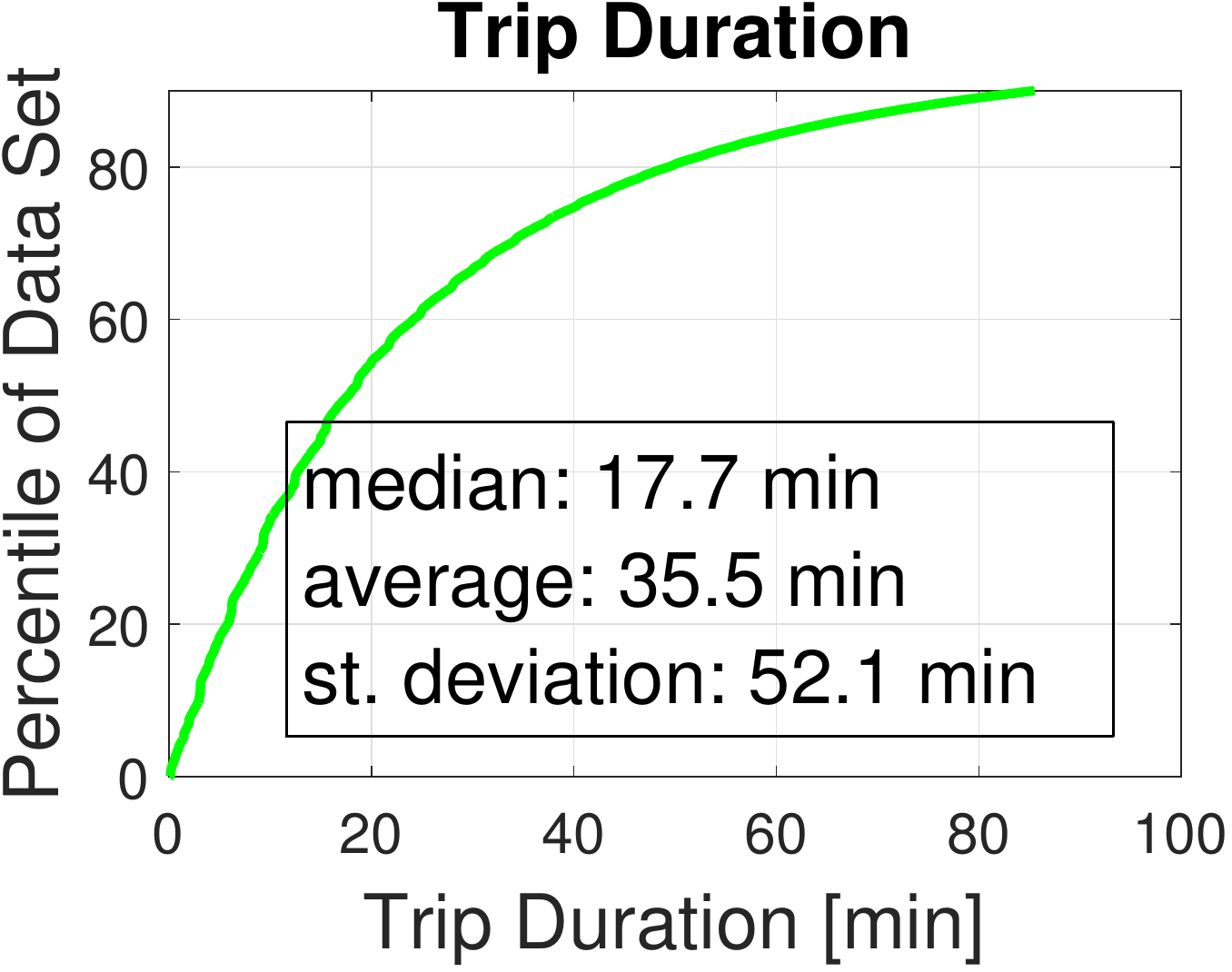}
	\end{subfigure}%
	\begin{subfigure}[b]{0.16\textwidth}		
		\includegraphics[height=24mm]{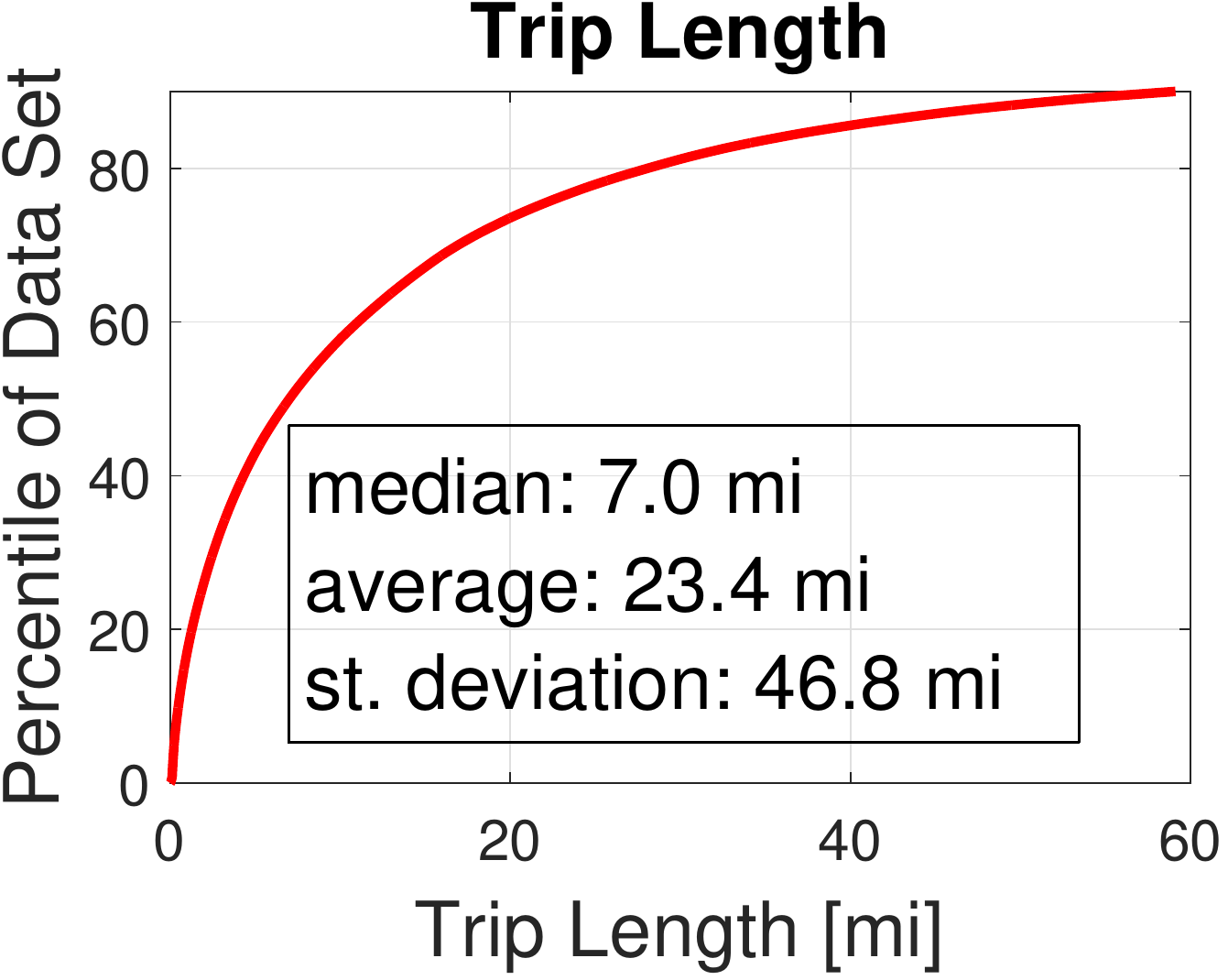}
	\end{subfigure}
	\caption{A sample trip with relatively few waypoints and descriptive statistics for 6.4 million trips recorded in October. Trip lengths are computed based on great-circle distances between waypoints.}\label{SampleTripAndTripStat}
\end{figure}

\begin{figure}	
	\begin{subfigure}[b]{0.16\textwidth}
		\centering
		\frame{\includegraphics[height=24mm]{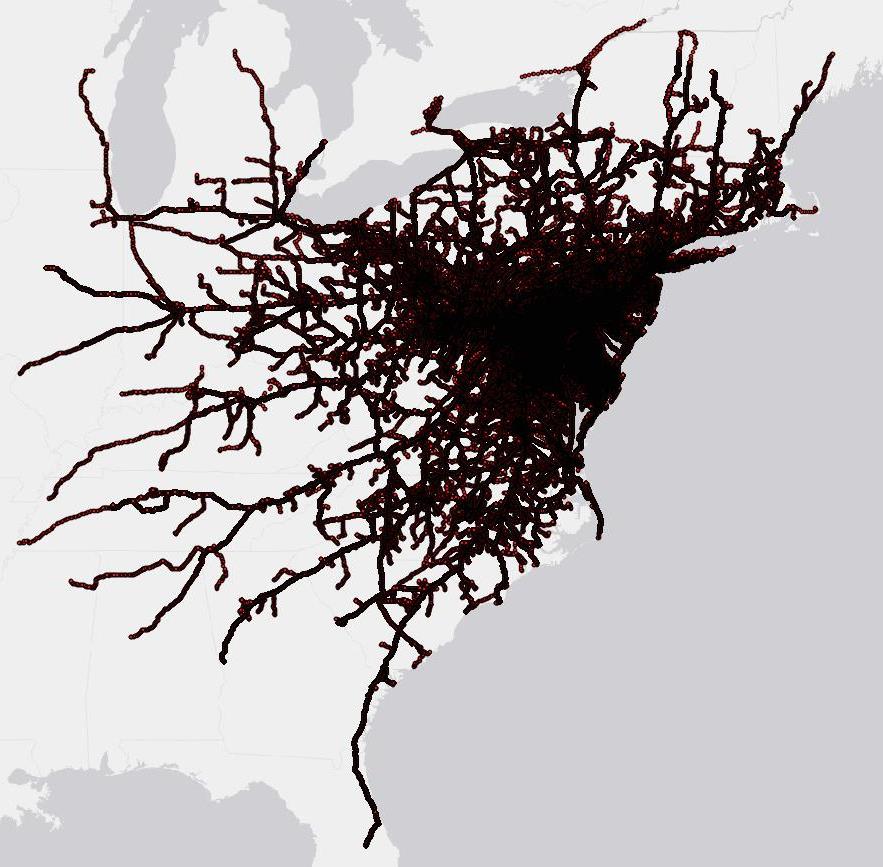}}
	\end{subfigure}%
	\begin{subfigure}[b]{0.16\textwidth}
		\centering
		\includegraphics[height=24mm]{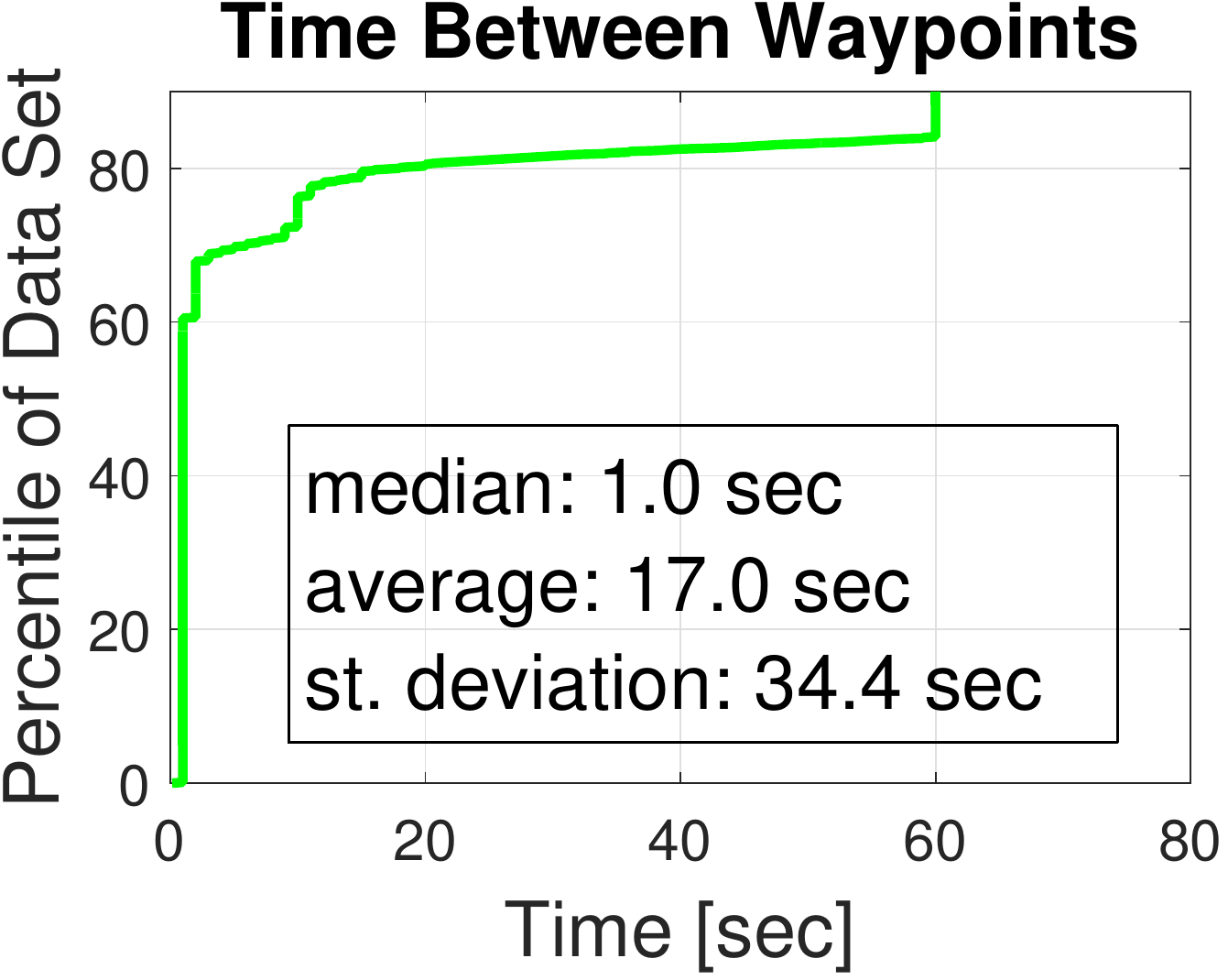}
	\end{subfigure}%
	\begin{subfigure}[b]{0.16\textwidth}
		\includegraphics[height=24mm]{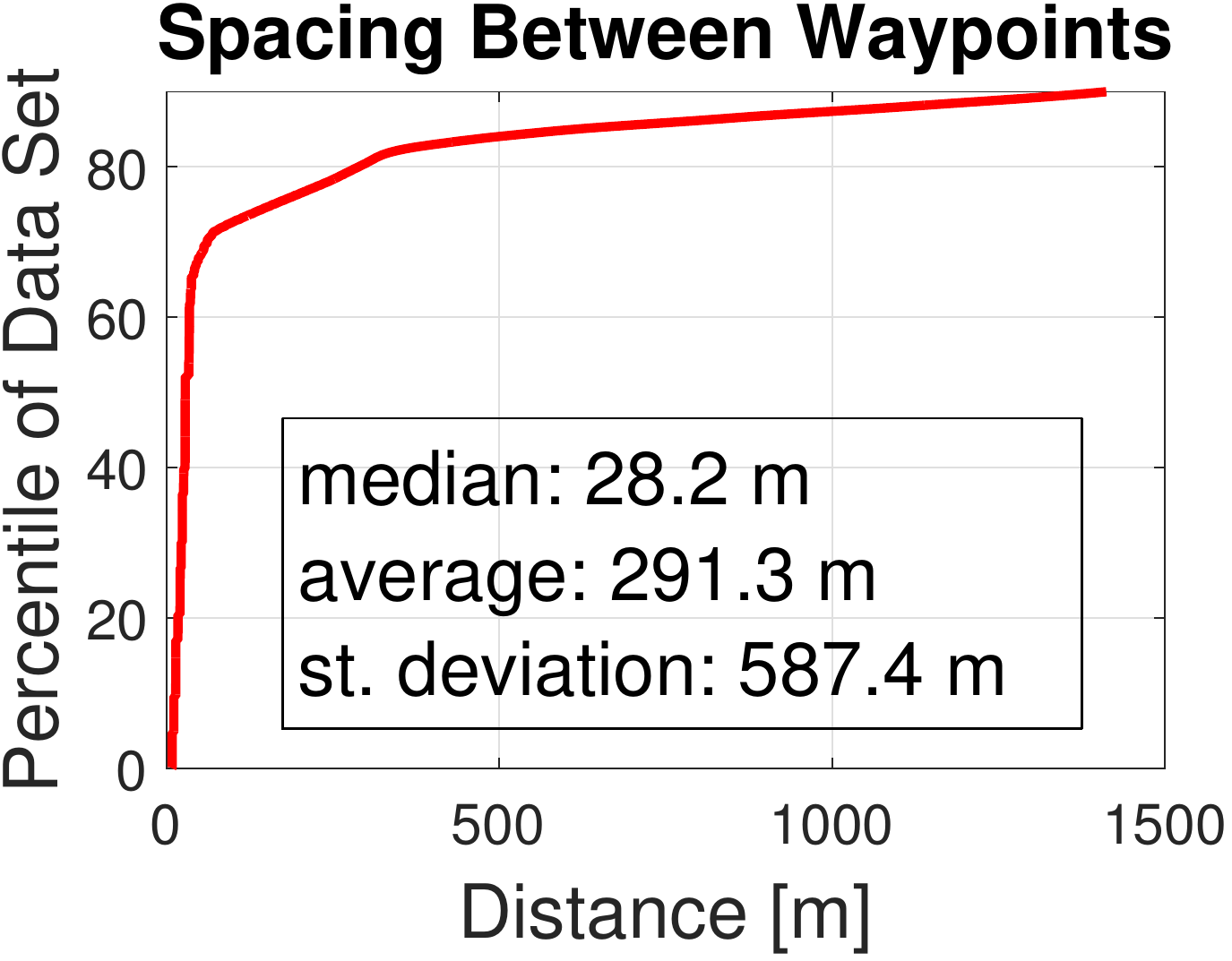}
	\end{subfigure}
	\caption{October waypoints and statistics computed based on a sample of over 360 million waypoints after removing outliers (e.g., unrealistic displacements due to device-related errors). Spacing is expressed in great-circle distances.}\label{WaypointsStat}
\end{figure}

\begin{figure}
	\centering
	\scalebox{.55}{
		\includegraphics[height=40mm]{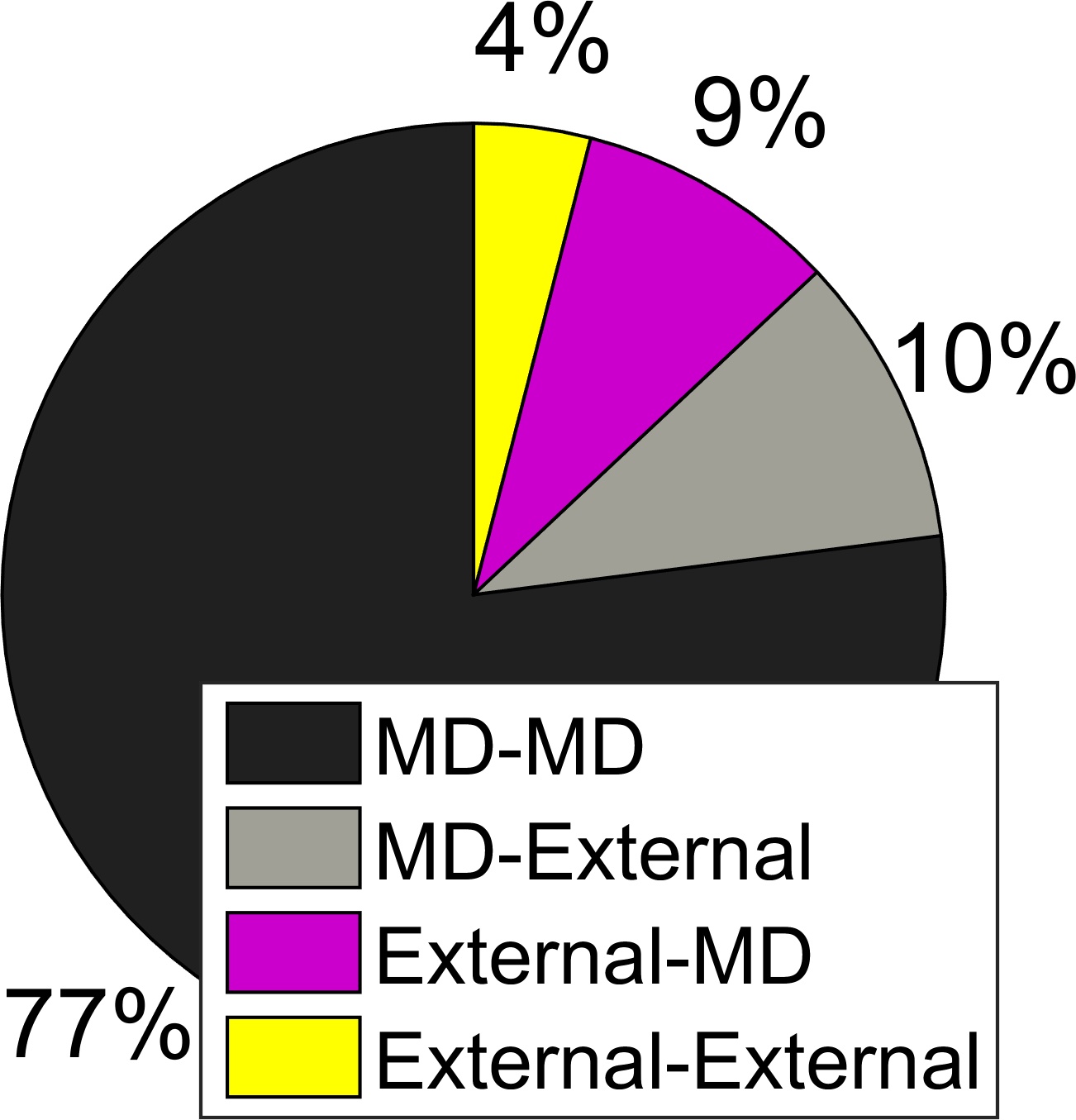}\includegraphics[height=40mm]{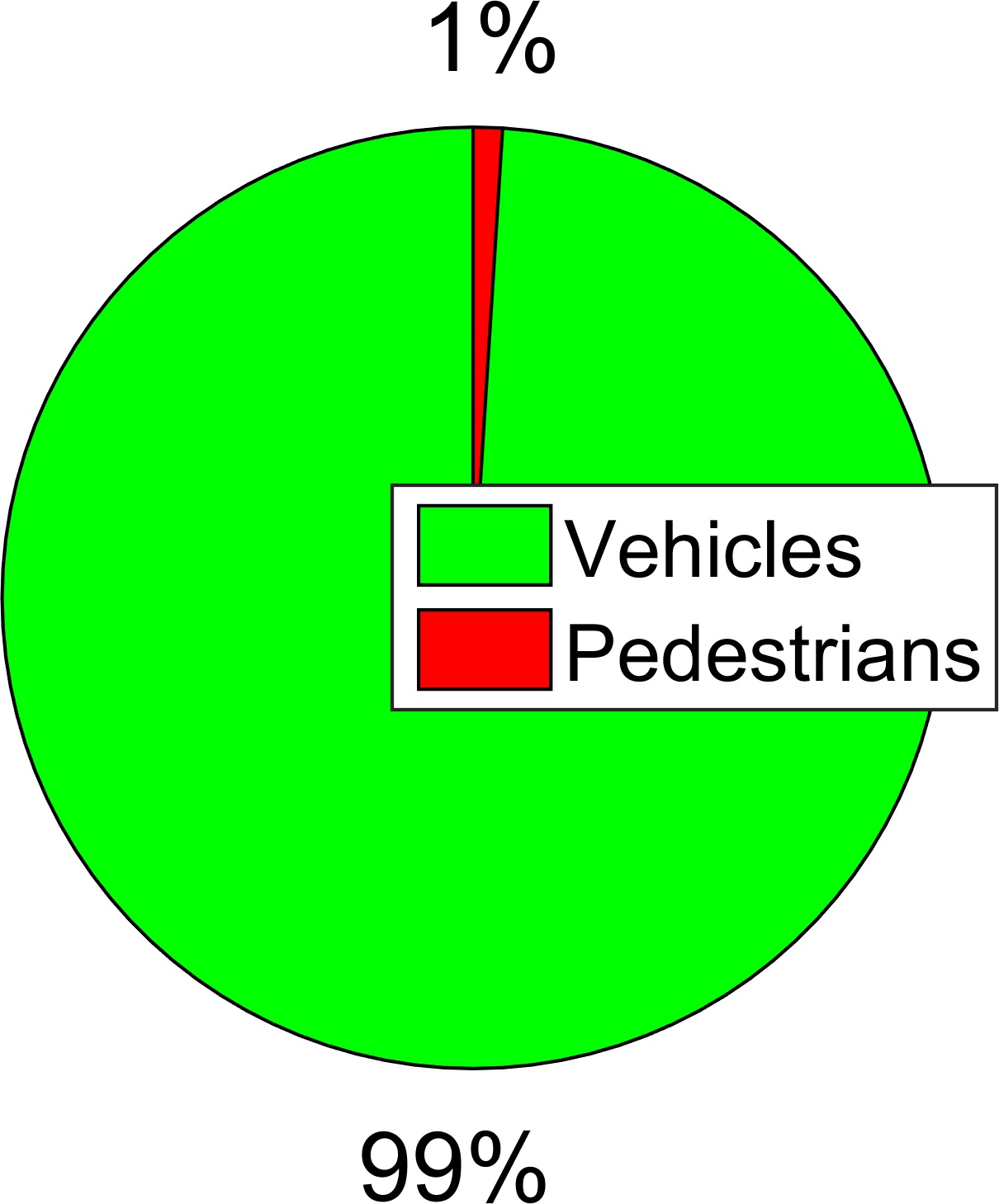}\includegraphics[height=38mm]{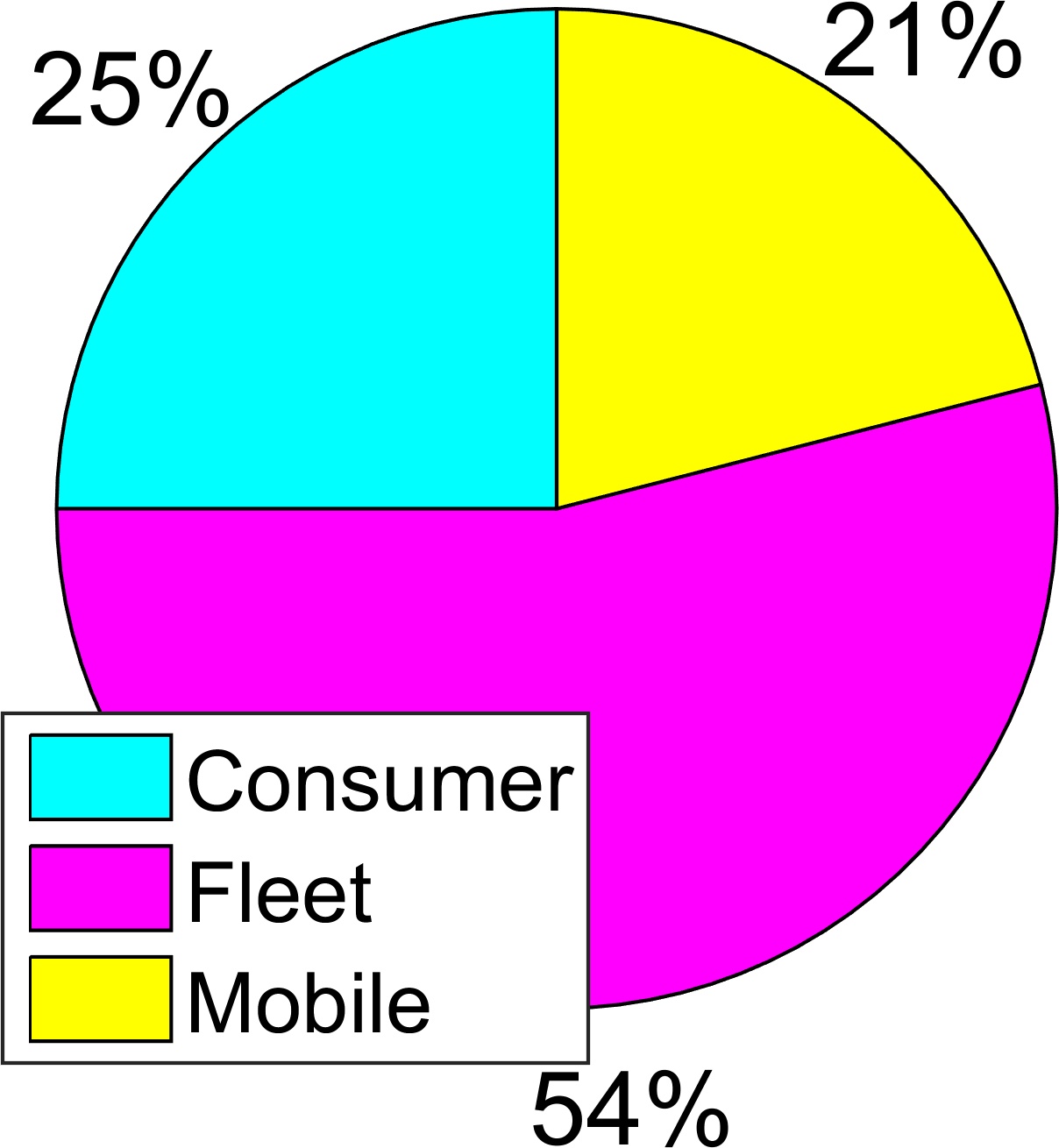}\includegraphics[height=38mm]{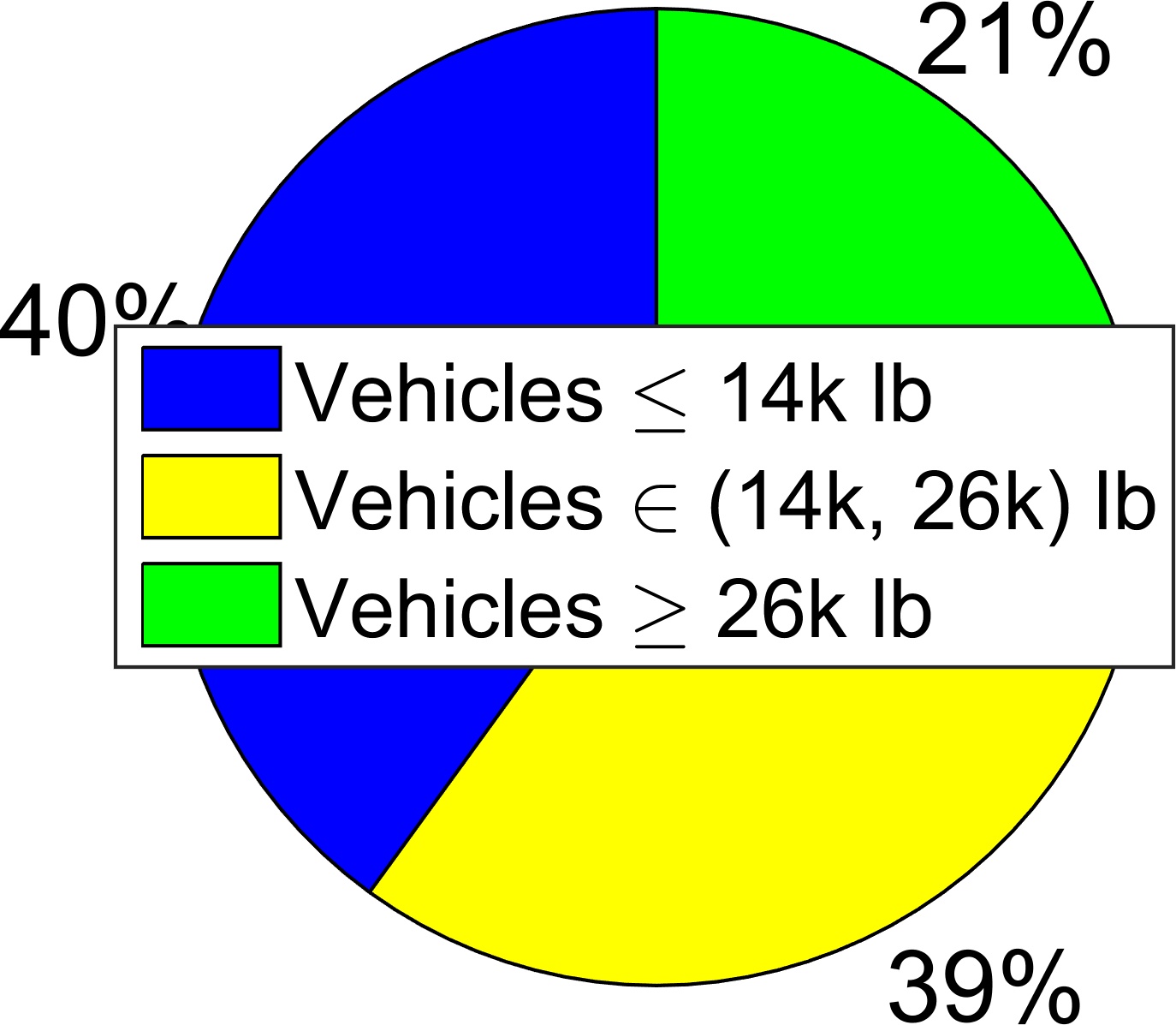}
	}
	\caption{Summary of trip attributes for 6.4 million October trips: geospatial, mode, provider type and vehicle weight classes.}\label{TripAttributes}
\end{figure}

The dataset used in this paper consists of GPS trajectories from 20 million trips recorded during February, June, July and October of 2015. Each trip consists of an origin and destination, as well as a number of intermediate waypoints, each of which has a corresponding time stamp (see Figure \ref{SampleTripAndTripStat} for a sample trip). Insight into the dataset is provided by summarizing characteristics of trips recorded during the month of October. Namely, the median trip duration and length are about 18 min and 7 miles (Figure \ref{SampleTripAndTripStat}), while the median time lapse and spacing between consecutive waypoints are approximately 1 second and 28 meters respectively (Figure \ref{WaypointsStat}). About 77\% of the trips are internal to Maryland, while the remaining 23\% have at least one waypoint outside Maryland (Figure \ref{TripAttributes}). The same visual indicates that the vast majority of trips correspond to vehicles (which are subdivided into three weight classes) while about 1\% of all the trips are pedestrian movements. In addition, Figure \ref{TripAttributes} shows that most trips pertain to fleet vehicles. In total, the raw GPS traces include 1.4 billion waypoints which requires 112 GB of storage space.

\begin{figure}
	\centering
	\begin{subfigure}[b]{0.23\textwidth}
		\centering		
		\frame{\includegraphics[width=\textwidth]{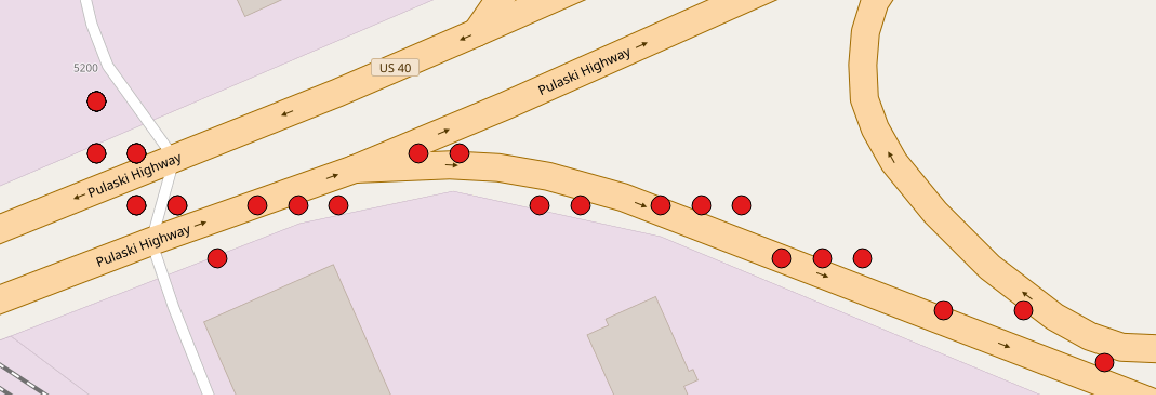}}
	\end{subfigure}\hfill
	\begin{subfigure}[b]{0.23\textwidth}
		\centering
		\frame{\includegraphics[width=\textwidth]{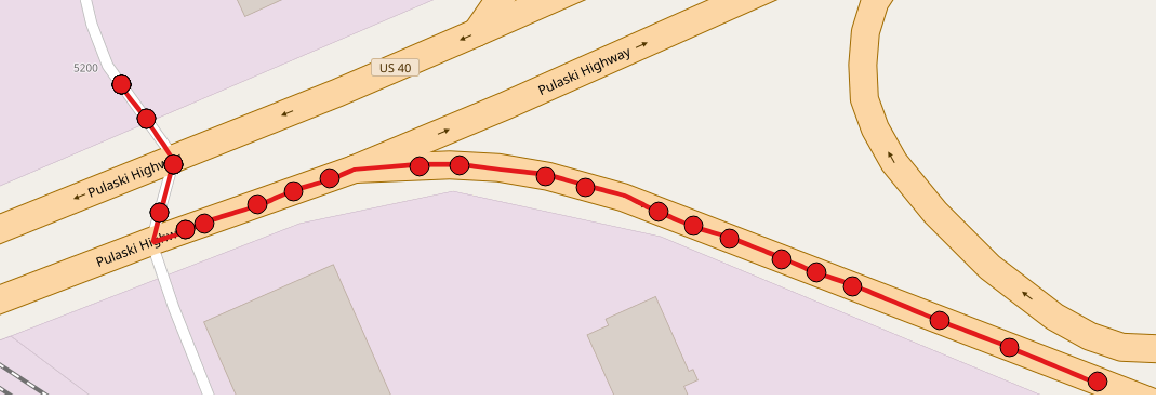}}
	\end{subfigure}	
	\caption{OpenStreetMap routing tool is applied to map match time-stamped sequence of latitude/longitude pairs (left) to most likely road-based routes (right).}\label{fig:OpenStreetMap}
	\label{fig:osm-routing}
\end{figure} 

\subsubsection{Preprocessing}\label{sec:mapmatching}
Since GPS data includes measurement errors, the recorded waypoints are not necessarily located along the physical road network. In addition, the granularity of data is not always high enough to include a waypoint along every single road link (or a traffic message channel) that a vehicle traverses. Therefore some preprocessing is needed in order to map match waypoints to the road network and reconstruct road-based routes. This was done using the OpenStreetMap \cite{haklay2008openstreetmap} routing tool (Figure \ref{fig:osm-routing}), which applies a hidden Markov model to find the most likely road-based route from a time-stamped sequence of latitude/longitude pairs \cite{newson2009hidden}. The computationally-intensive map matching was carried out in parallel on a 10-core computer, and took about 3 days to process all 20 million trips. Since map matching results in trajectories that include a significant amount of additional information (i.e., data about every road link that a vehicle traverses), the corresponding dataset increased in size from the initial 112 GB to over 5 TB. However, after removing redundant information (i.e., keeping only one node per road link), the remaining dataset was reduced from 5 TB to 700 GB. 

\subsubsection{Database}
In order to efficiently store and query the large dataset we utilized PostgreSQL 9.6, an open-source database that has several useful features for analyzing spatio-temporal data. 
First, it comes with PostGIS spatial database extender, which adds support for geographic objects and allows location queries to be run in the Structured Query Language (SQL). It is also highly integrated with QGIS, which is an open source Geographic Information System (GIS) that was extensively used in this study. Additionally, it includes a number of built-in solutions to facilitate data manipulations, such as table inheritance mechanism, spatial indexing, and advanced spatial queries. Finally, it is widely-used for processing spatial data, which results in a sizable online community and support. The primary disadvantage of using PostgreSQL is the limitation regarding parallel queries (i.e., the planner will not conduct a parallel query if it involves any data writing). However, this limitation will likely be removed in future releases of PostgreSQL.

\subsubsection{Penetration rate}\label{PRsection}
\begin{figure}
	\centering
	\frame{\includegraphics[height=40mm]{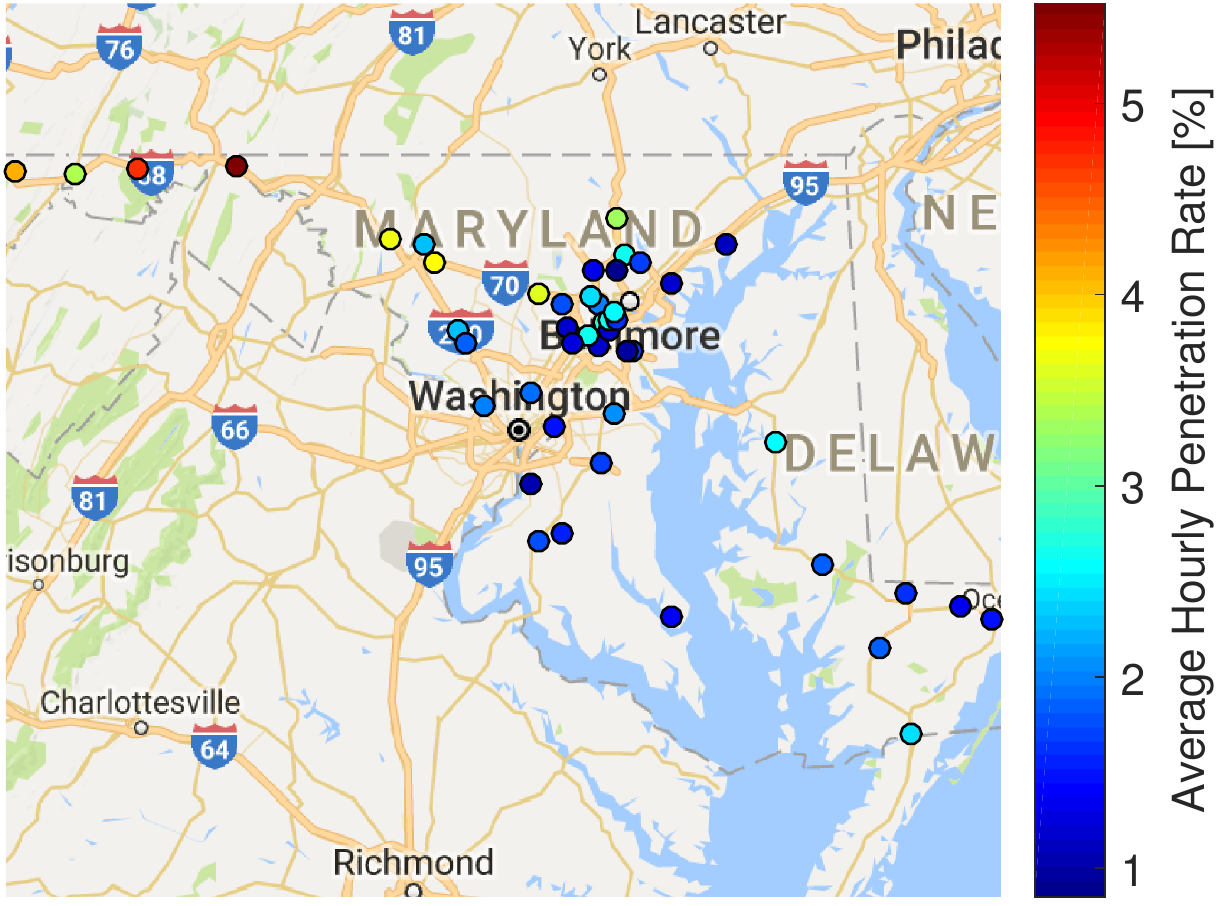} \, \includegraphics[height=40mm]{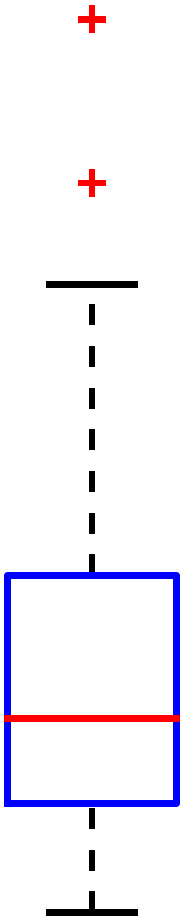}}
	\caption{Penetration rates of recorded trips are estimated via comparison with over 224,000 hourly records from 47 ATR stations. The average hourly PRs vary over these 47 locations from 0.85\% to 5.52\%, with the median of 1.86\%.}\label{ATRcomparison}
\end{figure}

Because the trajectory data represents only a subset of vehicles on the road, it is important to roughly quantify the penetration rate (PR) of the analyzed trips. Doing so may help indicate the extent to which the sample is representative of overall traffic, and also provide insight into the total number of vehicles traveling on road segments between fixed traffic sensors. To perform rough PR estimates, we compared GPS traces and data from 47 automatic traffic recorder (ATR) stations in Maryland, which typically provide hourly vehicle counts without differentiating between vehicle types. The average hourly PRs at 47 locations are provided in Figure \ref{ATRcomparison}, which indicates that average PRs at these 47 locations vary from 0.85\% to 5.52\%, with a median of 1.86\%. This implies that observed trips capture one in every 54 vehicles.

\subsection{Methods}\label{sec:MethodDescription}
We employ an array of machine learning algorithms and data visualization techniques to extract value from 20 million GPS traces and effectively communicate our results with transportation agencies. Here we provide an overview of the clustering algorithms used in the analysis, as well as software solutions that the authors found particularly useful in analyzing and visualizing trajectory data. This discussion should provide a brief guideline to transportation agencies that are beginning to analyze trajectory data.

\subsubsection{Density-based clustering}	 
DBSCAN (density-based spatial clustering of applications with noise) is a widely-applied clustering algorithm \cite{ester1996density}, which identifies each data point as a core point, border point, or outlier based on two input parameters: $\varepsilon$ and \textsc{MinPts}. $\varepsilon$ is a radius parameter that defines the $\varepsilon$-neighborhood $N(\varepsilon)$ around each point, and \textsc{MinPts} represents the minimum number of data points in $N(\varepsilon)$ required to form a core point. Clusters are built around core points (which represent high-density areas) by iteratively adding density-connected points. DBSCAN does not require the number of clusters as an input parameter, can easily find arbitrarily-shaped clusters, is robust with respect to outliers (which are treated as noise and do not affect existing clusters), and is also implemented in many libraries which facilitates its application. However, one of the disadvantages is that it is very sensitive to input parameters \cite{karypis1999chameleon}, where small changes to the radius and distance parameters can yield different clustering results. In addition, the definition of distance should be carefully considered because it naturally affects the results (e.g., see \cite{pelekis2012visually} for a related discussion of similarity measures for trajectories). This paper utilizes DBSCAN for constructing isochrones based on trajectory data. Finally, a related density-based clustering algorithm OPTICS (ordering points to identify the clustering structure) \cite{Ankerst1999optics} is used in other applications, as described later in the paper.

\subsubsection{Software}
V-Analytics (formerly Descartes and CommonGIS) is a free visual data exploration and visual analytics software that facilitate exploration, analysis and modeling of different kinds of spatio-temporal data: events, time series, trajectories and situations. The system includes a variety of interactive visualization techniques \cite{andrienko2006exploratory}, supports necessary transformations of spatio-temporal data \cite{andrienko2013visual} and integrates a number of computational methods, adapted for analysis in space and time. Particularly, tools for clustering trajectory data with a library of suitable similarity measures are integrated \cite{andrienko2009interactive}. We used V-Analytics to do quick exploratory analysis, compare performance of different clustering algorithms, and obtain high-quality visuals.

QGIS is an open-source GIS tool developed through the Open Source Geospatial Foundation \cite{QGIS_software}. QGIS was used to prepare the majority of maps and map-based animations in this work. We found QGIS particularly useful due to its interface with PostgreSQL for easy preparation and management of large datasets, its interface with Python for programmatic manipulation of maps and their appearance, and the large online community that provides support and numerous plug-ins written in Python and C++.

\subsection{O-D matrices}
\begin{figure*}	
	\begin{subfigure}{0.33\textwidth}
		\centering
		\includegraphics[height=60mm]{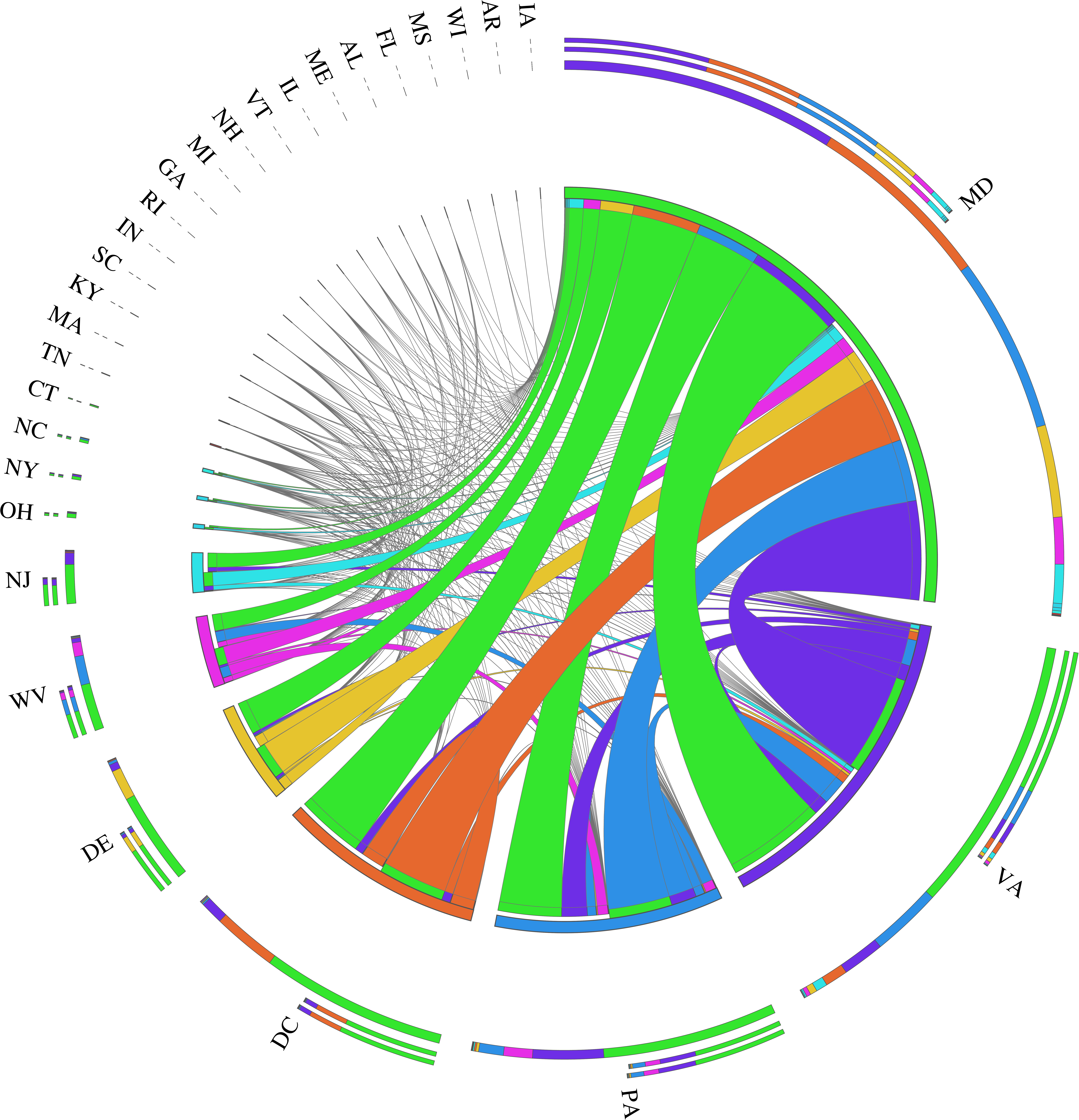}
		\caption{Trips between MD and other states}\label{CircosPlot1}
	\end{subfigure}%
	\quad
	\begin{subfigure}{0.33\textwidth}
		\centering
		\includegraphics[height=60mm]{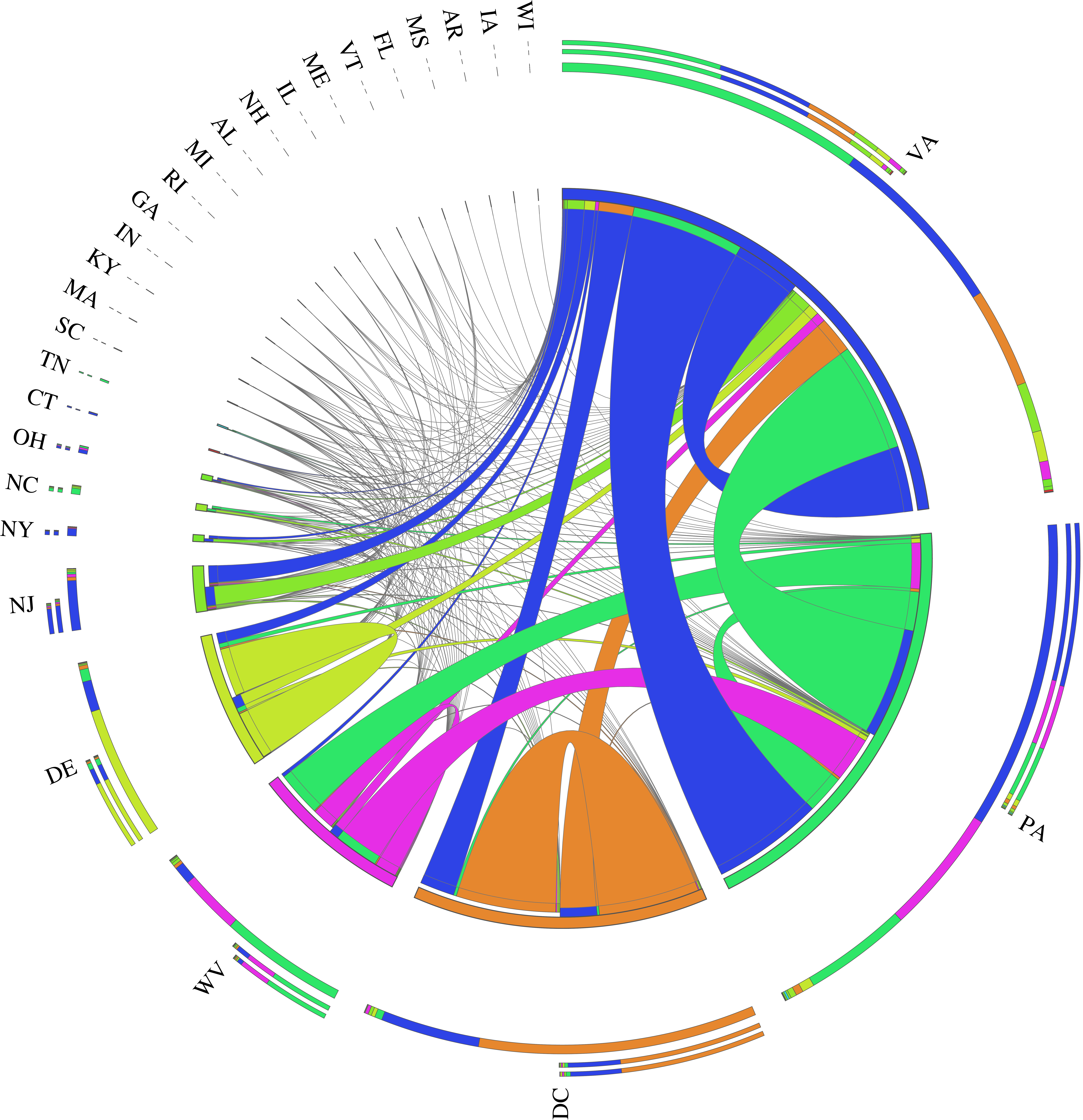}
		\caption{Trips traversing MD}\label{CircosPlot2}
	\end{subfigure}
	\quad
	\begin{subfigure}{0.33\textwidth}
		\centering
		\includegraphics[height=60mm]{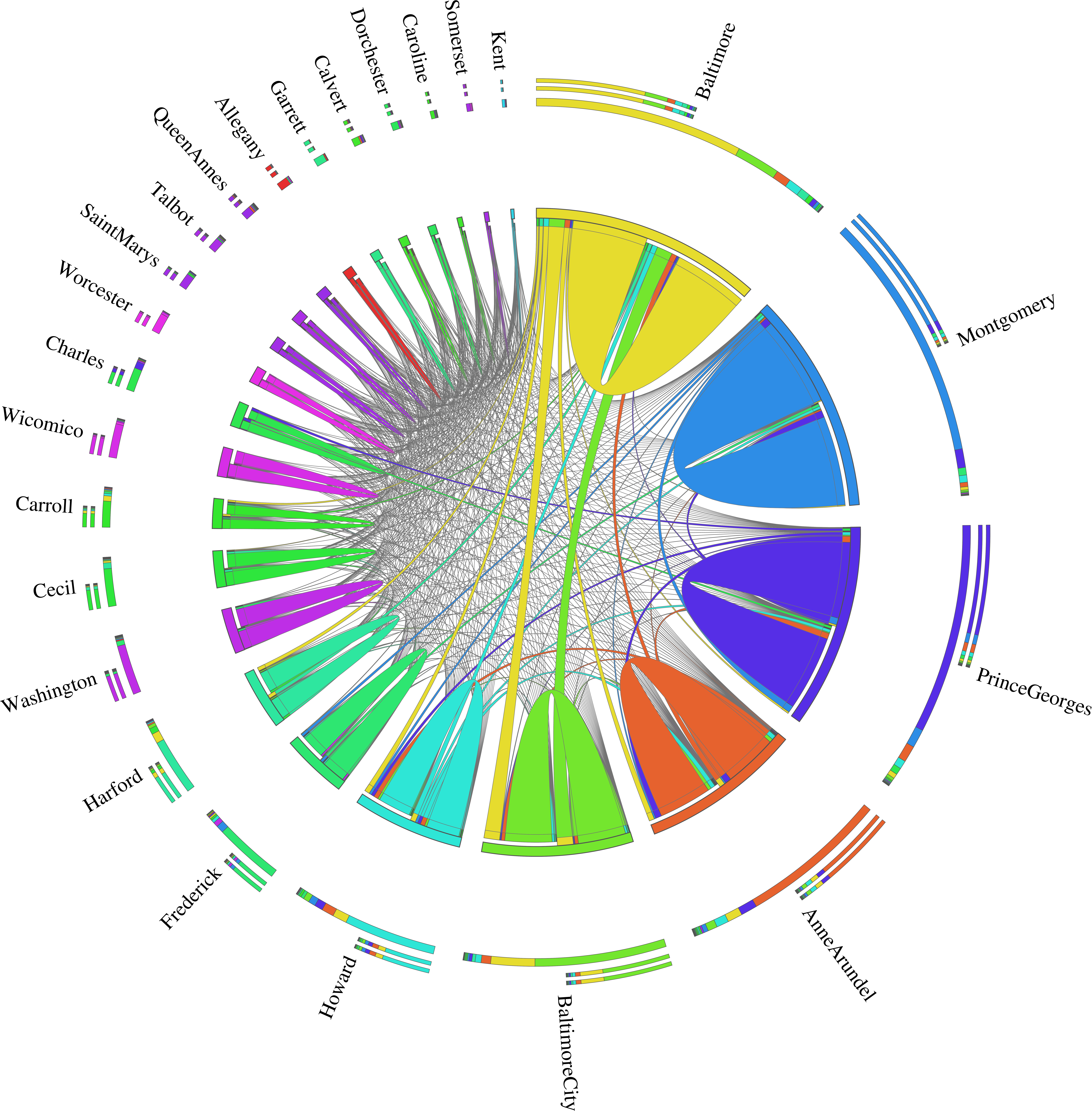}
		\caption{Trips within and between counties in MD}\label{CircosPlot3}
	\end{subfigure}%
	\caption{O-D matrices visualized with Circos \cite{krzywinski2009circos}.}\label{CircosPlots}
\end{figure*}

As argued in the literature review, demand modeling and transportation planning relies on estimating the number of trips that take place between specific locations \cite{Iqbal2014}. To illustrate the value of trajectory data in estimating demand, we map the origins and destinations of the 20 million trips to geographic areas of different sizes (i.e., traffic analysis zones, zip codes, counties and states), and visually explore the corresponding O-D matrices. While dense O-D matrices are somewhat difficult to visualize, those with fewer entries can be visually explored using open-source software Circos \cite{krzywinski2009circos}. For example, Figure \ref{CircosPlot1} depicts GPS trips between Maryland and other states, where the green ribbons denote trips originating in Maryland and ending in other states. This visual indicates that most trips originate and end in few neighboring states (i.e., Virginia, Pennsylvania), which are ordered clock-wise based on the total number of trips. It also shows that the number of trips going in and out of Maryland is balanced, which can be observed by comparing the two outermost concentric circles that are of approximately same length and color pattern. Figure \ref{CircosPlot2} visualizes the subset of these trips that traverse Maryland, and indicates that a notable number of trips that originate and end in a neighboring state (e.g., District of Columbia, Delaware, Virginia) still use the Maryland infrastructure. Figure \ref{CircosPlot3} shows a county-based O-D matrix for trips internal to MD and suggests that most trips originate and end within the same county, which is an expected result because the median trip length is about 7 miles (Figure \ref{SampleTripAndTripStat}). As previously mentioned, we can also map GPS trips to smaller areas (e.g., zip codes and traffic analysis zones), and explore the corresponding O-D matrices via interactive applications (e.g., GIS, web). 

Since the analyzed GPS traces represent only a sample of all vehicles on the road, the O-D matrices shown in Figure \ref{CircosPlots} need to be scaled by appropriate expansion factor(s) to estimate actual traffic. A rough estimate of the total number of trips between regions can be obtained by scaling O-D matrices in Figure \ref{CircosPlots} by the factor of 54 (see Section \ref{PRsection}), which is the approach that the authors of this paper will take to obtain an aggregate trip table needed as an input for a statewide transportation planning model. However, one could improve on this by trying to derive custom expansion factors for different O-D pairs, days of the week, and hours of the day. This analysis could be further improved by deriving O-D matrices for different types of vehicles (i.e., passenger cars vs. trucks); however, this would require determining PR of trajectory data for different types of vehicles (see Figure \ref{TripAttributes}), which would be possible with traffic sensors that can differentiate vehicle types. Unfortunately, this is not the case with Maryland ATR stations.
	
\begin{figure}
	\centering
	\begin{subfigure}[b]{0.35\textwidth}
		\centering		
		\frame{\includegraphics[height=62mm]{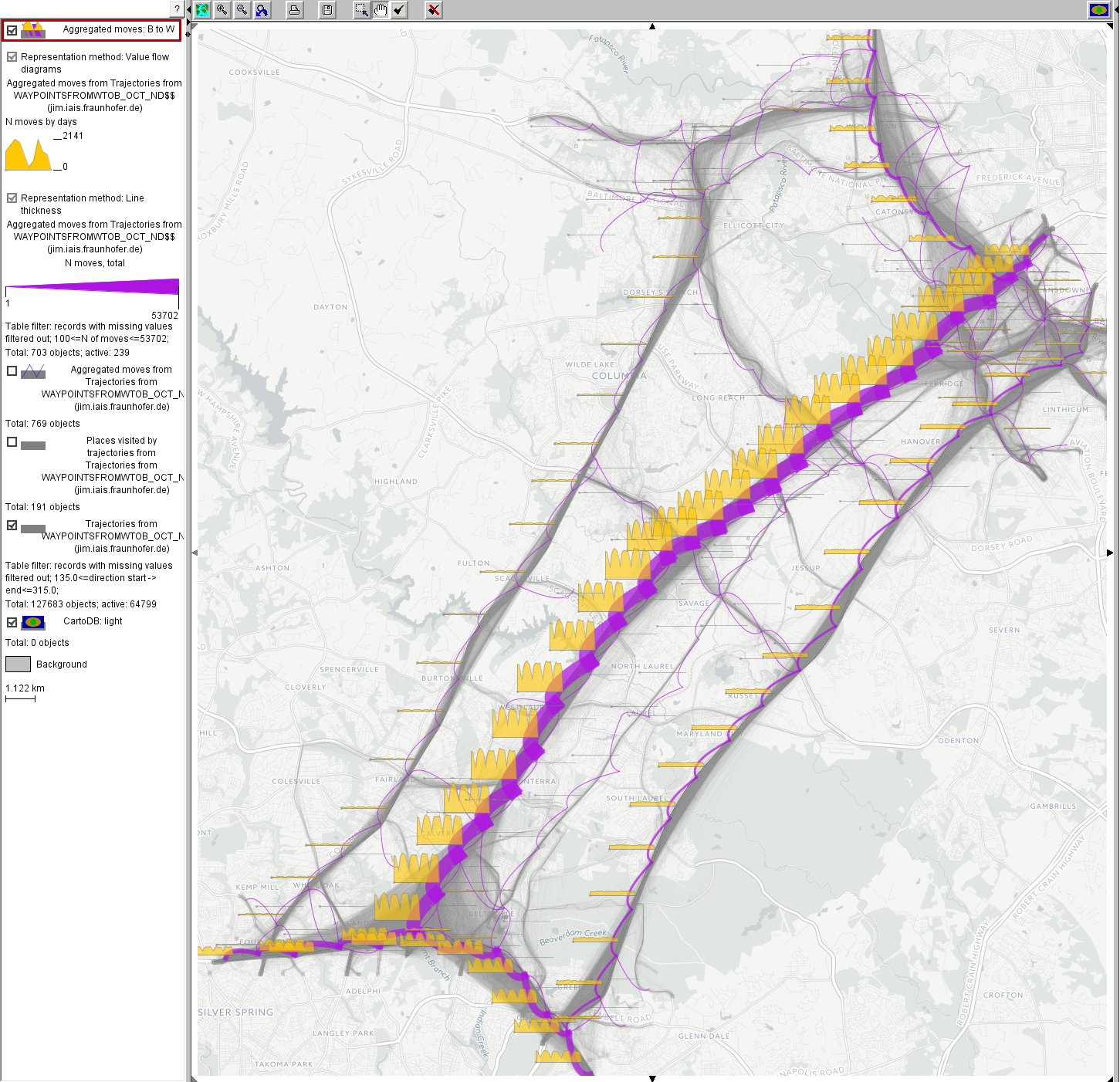}}
		\caption{Trips from Baltimore to Washington}\label{ICMtripsTrajectories}
	\end{subfigure}%
	~
	\begin{subfigure}[b]{0.13\textwidth}
		\centering
		\includegraphics[height=19mm]{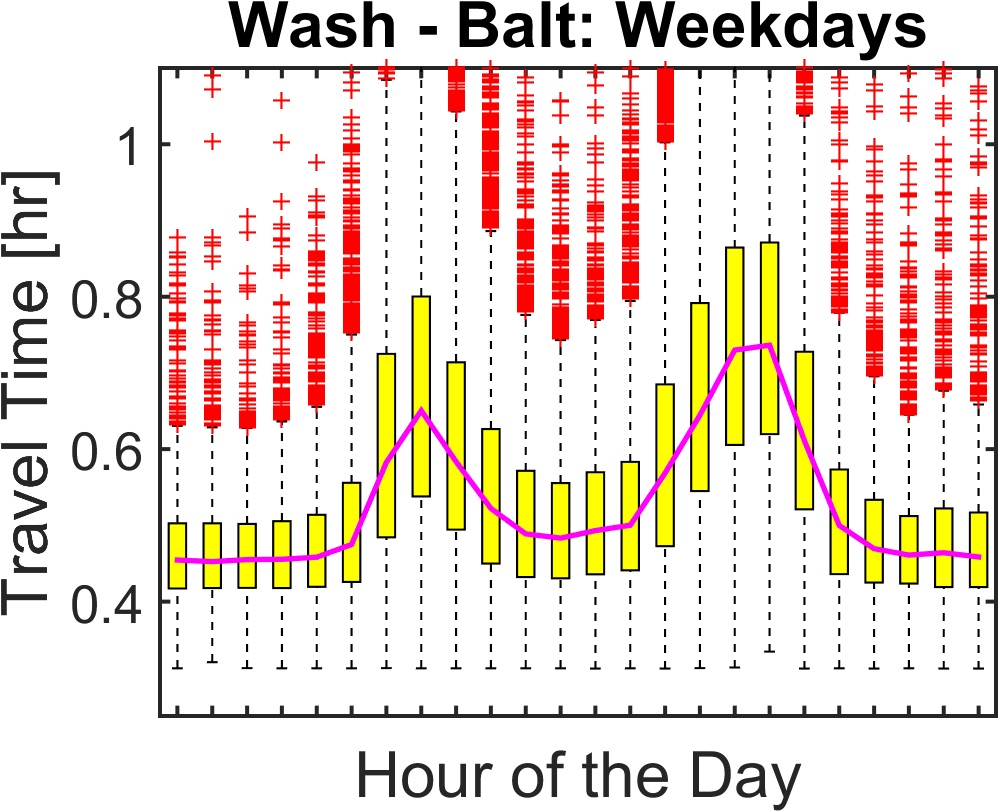} 
		
		\vspace{5pt}
		
		\includegraphics[height=19mm]{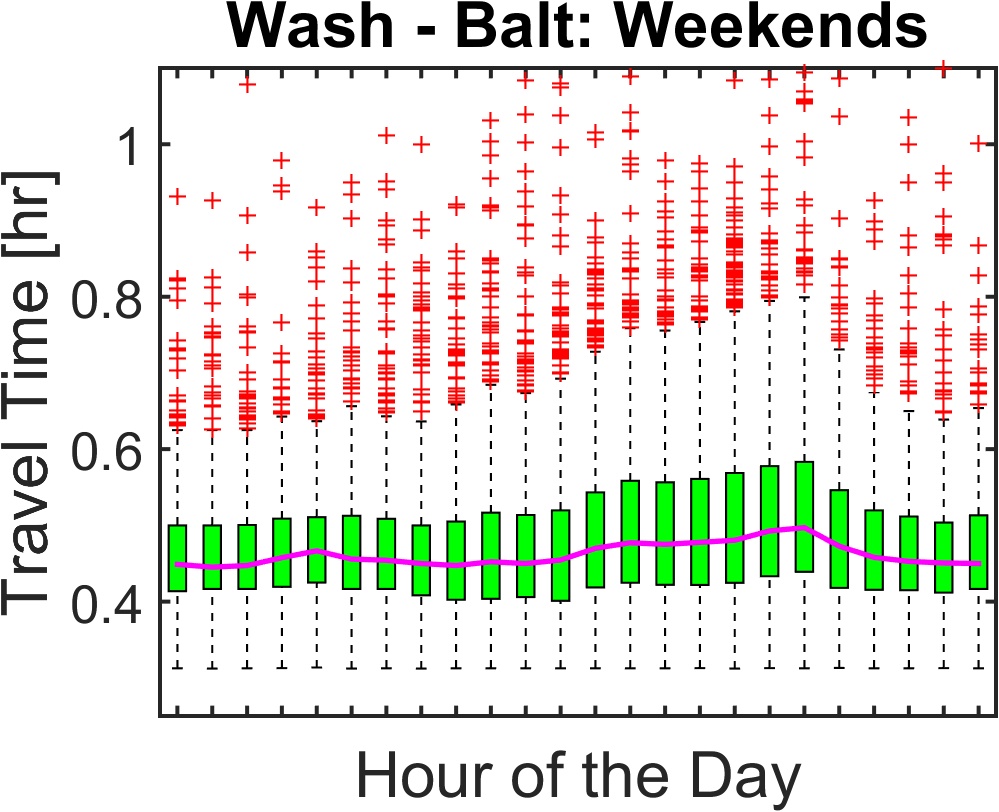} 
		
		\vspace{5pt}
		
		\includegraphics[height=19mm]{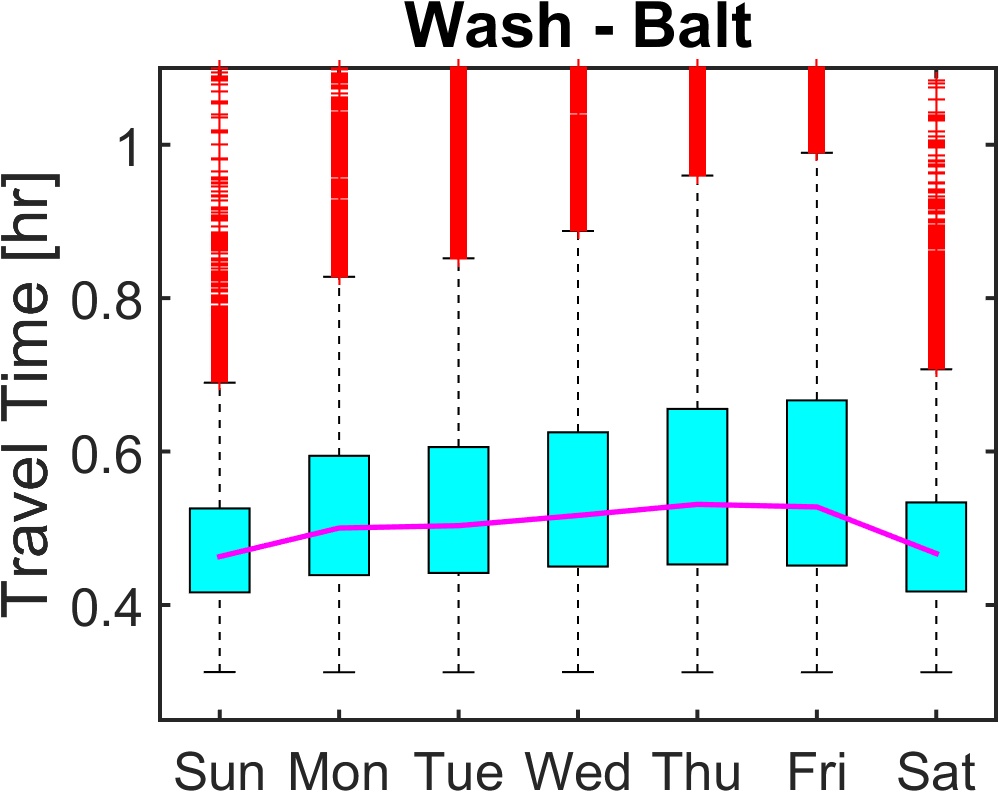}
		\caption{Trip statistics}\label{ICMtripsStatistics}
	\end{subfigure}
	\caption{Trajectories of trips between Washington and Baltimore beltways that took place during October. Boxplots show travel times for trips between the two beltways.}\label{ICMtrips}
\end{figure} 

\subsection{An O-D pair}
Rather than considering an entire O-D matrix at various levels of granularity, it is sometimes useful to focus on a specific O-D pair. To illustrate this, we consider trips between Washington and Baltimore (Figure \ref{ICMtrips}), and use GPS traces between this O-D pair to visually explore flow patterns, travel time variability and split rates amongst three major routes. Figure \ref{ICMtripsTrajectories} shows the raw trajectories as well as aggregated trips for days and links between neighboring polygons. Interestingly, both beltways and I-95 (the middle road) show clear weekly patterns, whereas I-295 (East-most road) has stable load with no weekly patterns. Moreover, Figure \ref{ICMtripsStatistics} visualizes travel times between the Washington and Baltimore beltways broken down by hour of day for weekday/weekend and day of week. On weekdays, the morning peak occurs for trips departing at 7-8 AM, while the afternoon peak is observed for trips departing at 4-5 and 5-6 PM. A very different travel pattern is observed on weekends, during which travel times are much steadier and also shorter than on weekdays. 

\subsection{Trip generators and isochrones}
\begin{figure}
	\centering
	\begin{subfigure}[b]{0.23\textwidth}
		\centering		
		\frame{\includegraphics[width=\textwidth]{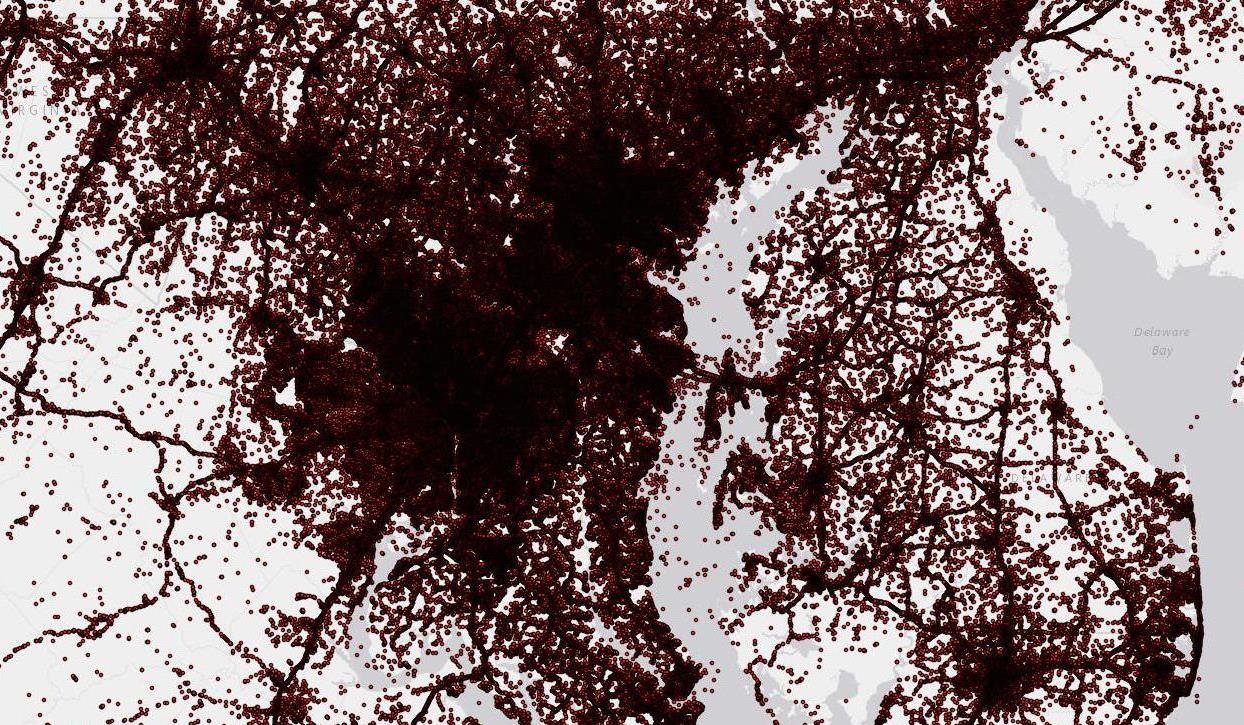}}
		\caption{Trip origins in MD}
	\end{subfigure}\hfill
	\begin{subfigure}[b]{0.23\textwidth}
		\centering
		\frame{\includegraphics[width=\textwidth]{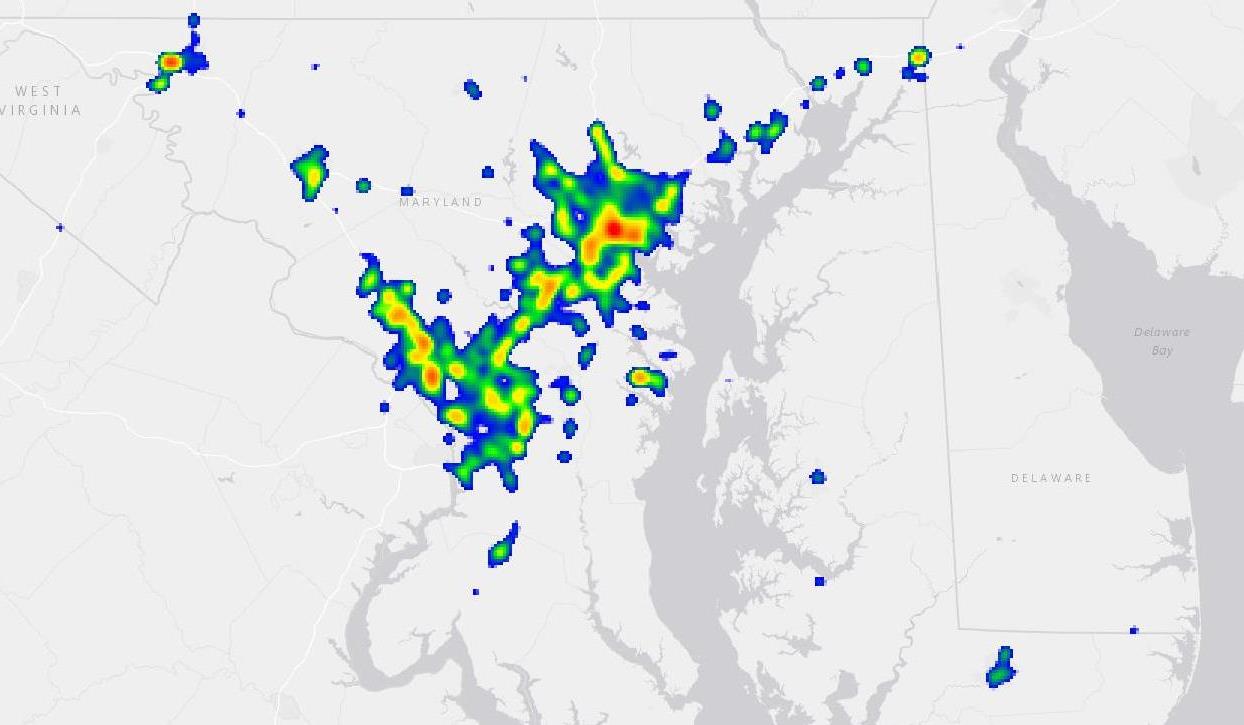}}
		\caption{Heat map of origins in MD}
	\end{subfigure}
	\caption{Some of the major trip generators: Baltimore downtown, BWI airport -- Fort Mead, Bethesda -- German Town.}\label{OriginLocations}
\end{figure}

In addition to analyzing GPS traces between O-D pairs, it is instructive to consider origins and destinations separately. For example, Figure \ref{OriginLocations} shows trip origins, which are spread over the entire state of Maryland. While the sheer number of data points obscures any patterns, creating and overlaying a simple heat map representing origin density shows that many of the trips originate at only a handful of locations. The main trip generators are downtown Baltimore, Baltimore-Washington International Airport, and the stretch between Bethesda and German Town. Upon identifying the main trip generators, we can query trips that originate in these areas and use their trajectories to construct isochrones. However, trajectory datasets often contain anomalous waypoints, which may skew mobility statistics and visualizations. Here we describe a density-based clustering approach that helps identify these outliers using the previously-described DBSCAN algorithm. 

\begin{figure}
	\centering
	\begin{subfigure}[b]{0.24\textwidth}
		\centering		
		\frame{\includegraphics[height=32mm]{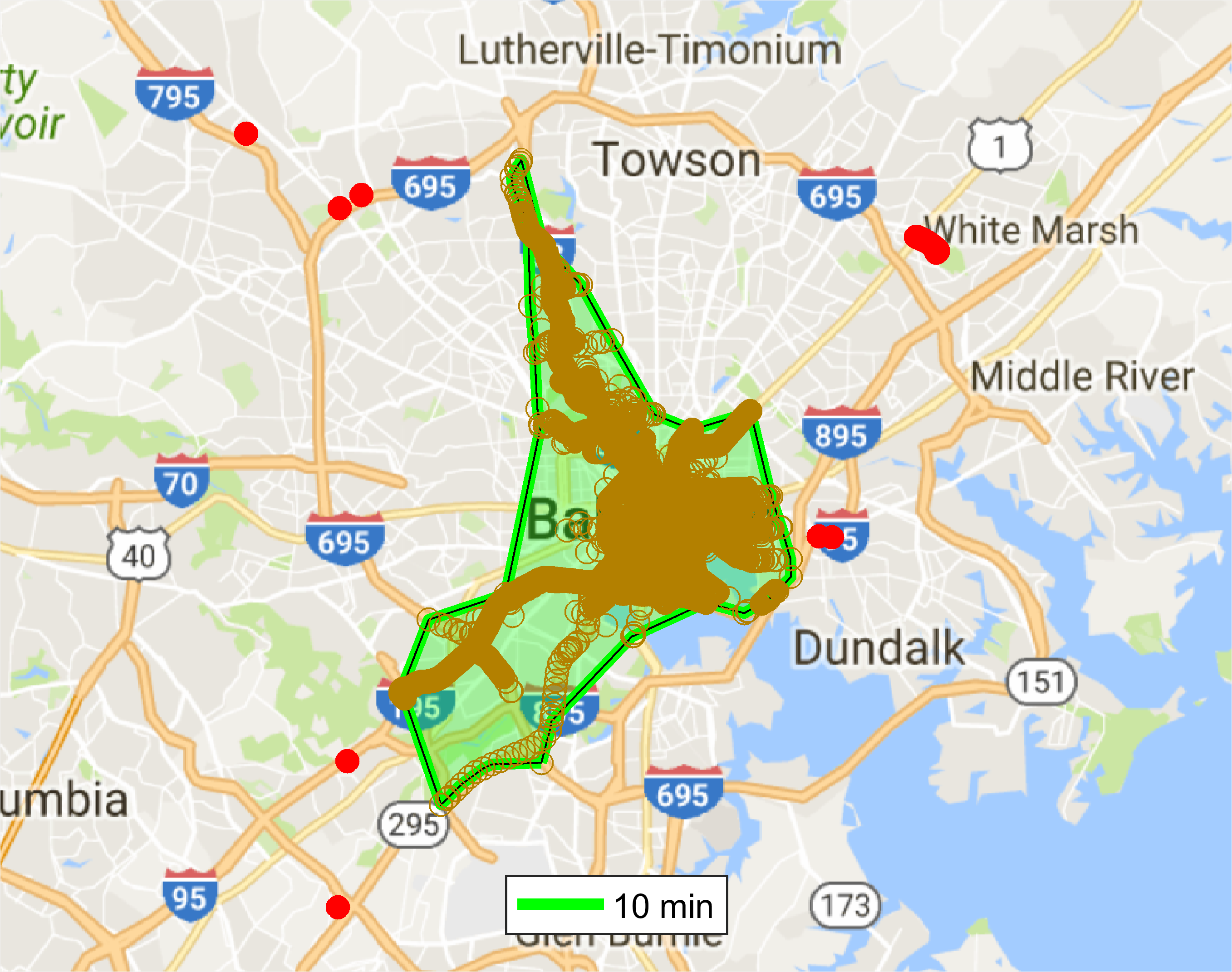}}
		\caption{10-min isochrone \& outliers}\label{fig:DBSCAN}
	\end{subfigure}%
	\begin{subfigure}[b]{0.24\textwidth}
		\centering
		\frame{\includegraphics[height=32mm]{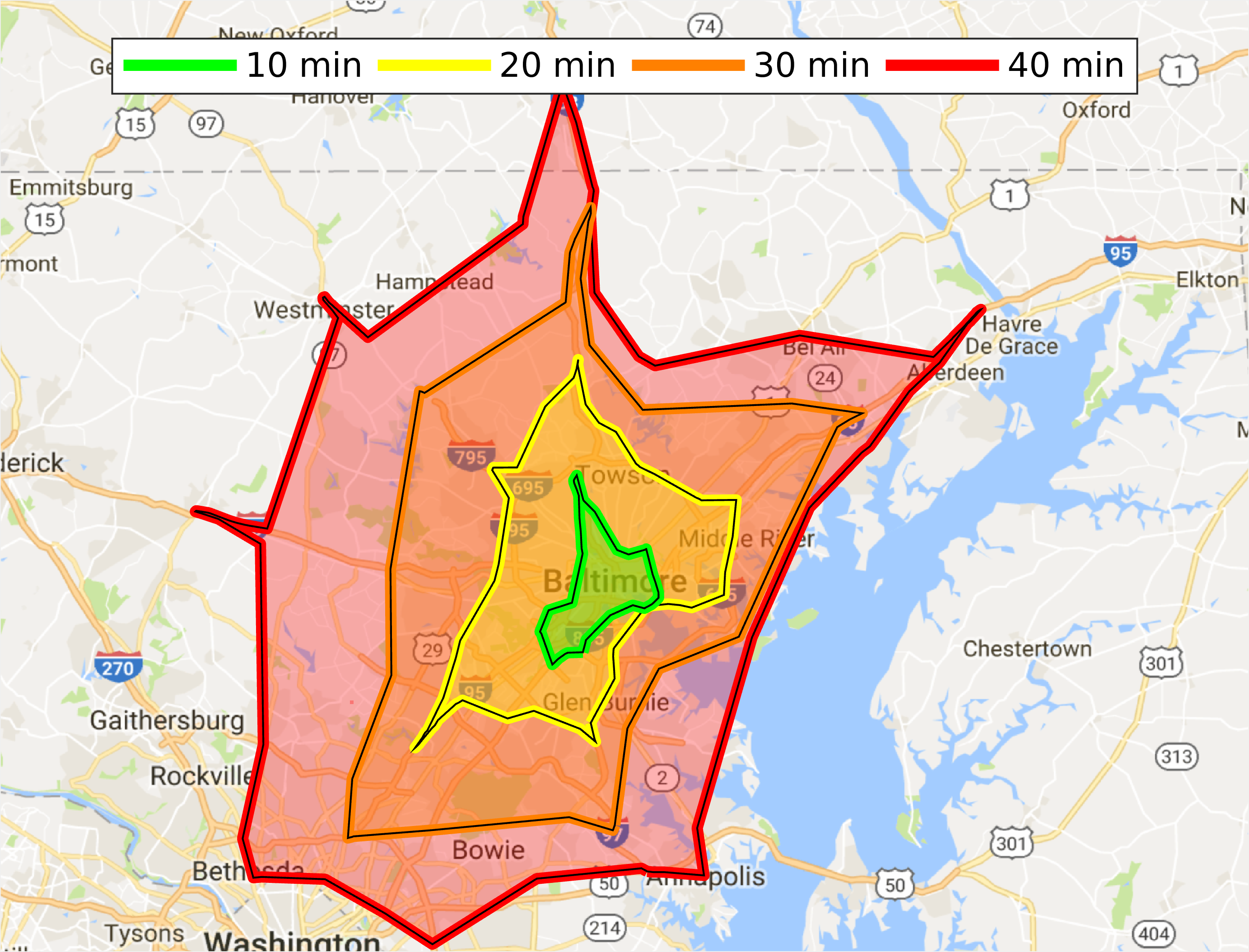}}
		\caption{Port of Baltimore isochrones}\label{fig:BaltimorePort}
	\end{subfigure}
	
	\begin{subfigure}[b]{0.2\textwidth}
		\centering
		{\renewcommand{\arraystretch}{1.2}
			\scriptsize{\begin{tabular}{ccc}\hline
					Isochrone	& $\varepsilon$ & \textsc{MinPts} \\
					(min)	& (km) & (pts) \\\hline
					10	& 1.1	& 60  \\
					20	& 1.3   & 20  \\ 
					30	& 1.4	& 10  \\
					40	& 1.6   & 5   \\ \hline
				\end{tabular}
		}}
		\caption{DBSCAN parameters}\label{DBSCANparameters}
	\end{subfigure}%	
	\begin{subfigure}[b]{0.28\textwidth}
		\centering
		\frame{\includegraphics[height=30mm]{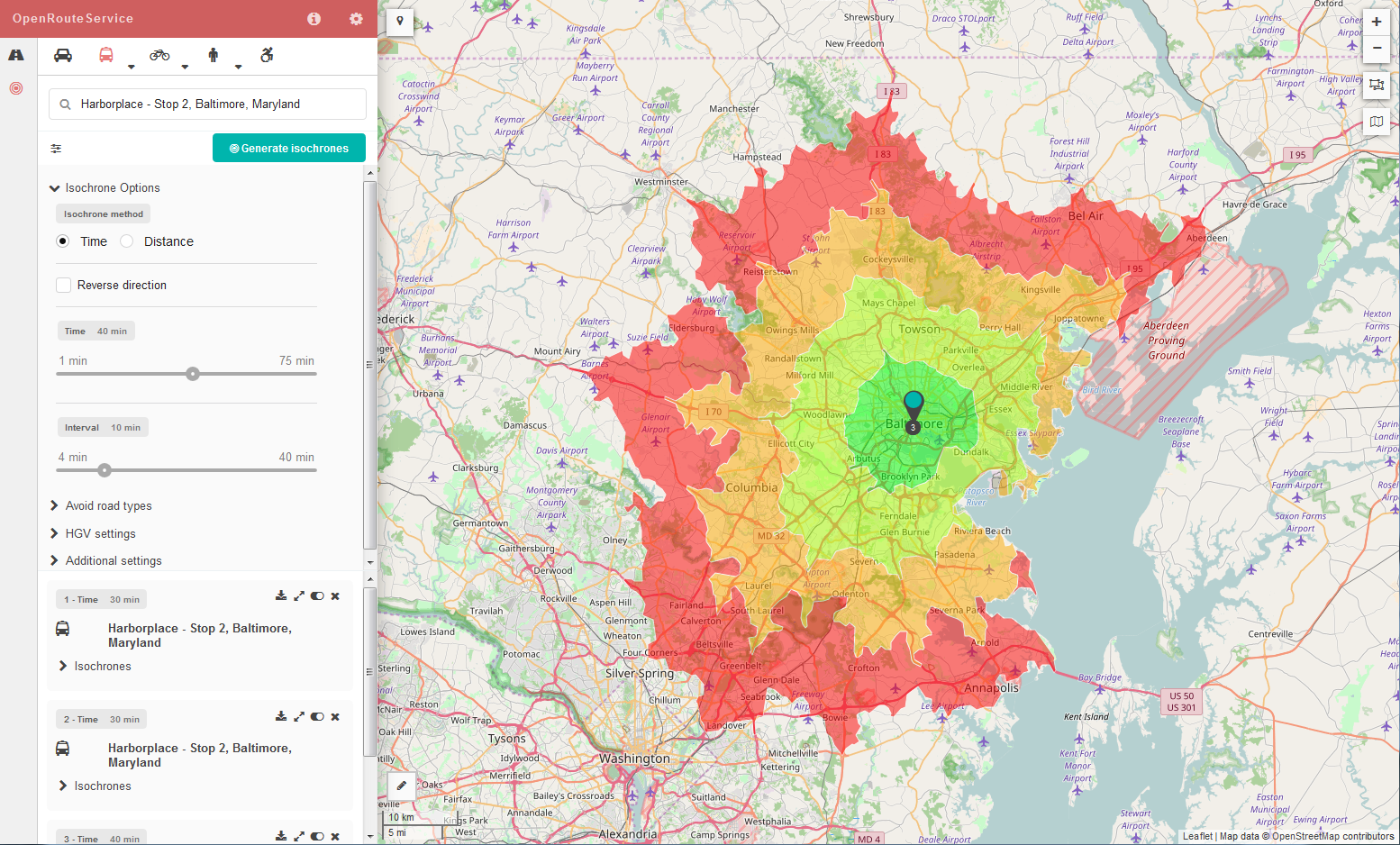}}
		\caption{Traditional isochrones}\label{TraditionalIsochrones}
	\end{subfigure}
	\caption{DBSCAN with the outlined parameters is used to construct isochrones from trip waypoints. After filtering waypoints based on density, the isochrone is obtained by constructing a concave hull which connects the boundary points (see Figure \ref{fig:DBSCAN} for an example). Traditional isochrones for heavy-vehicles from OpenRouteService \cite{neis2008openrouteservice} are used to validate the proposed clustering-based approach (compare Figures \ref{fig:BaltimorePort} and \ref{TraditionalIsochrones}).}\label{BWIisochrones} 
\end{figure}

To showcase this approach, we consider a set of trips originating from a single location and use the DBSCAN algorithm to identify outliers for 10, 20, 30, and 40 minute trips. As an example, we focus on a set of approximately 3,000 trips beginning from the Port of Baltimore, which consists of 218,302 total points (95,155 within 10 min, 141,586 within 20 min, 164,053 within 30 min, and 178,257 within 40 min). Using the scikit-learn Python implementation of DBSCAN \cite{pedregosa2011scikit}, we cluster the points for different combinations of input parameters (Figure \ref{DBSCANparameters}), remove the points that algorithm identifies as outliers, and visualize the results in the form of isochrones. Figure \ref{fig:DBSCAN} shows the results of running DBSCAN for points within 10 min travel time of the origin, with outlier points colored red, non-outlier points colored brown, and an isochrone defined as a single conforming 2-D boundary of non-outlier points. Note that, if the algorithm had not removed the marked outliers, the resulting concave hull would have included these points too, suggesting inflated levels of mobility. This procedure is repeated for 20, 30, and 40 minute trips, and the concave hulls bounding the non-outlier points are plotted in Figure \ref{fig:BaltimorePort}. The shape of the different concave hulls reflects the fact that mobility is greatest along the main highways, which matches our intuition. Finally, as a validation of the outlined approach, we note that isochrones for heavy vehicles designed based on a traditional method (Figure \ref{TraditionalIsochrones}) show very similar patterns to those observed in Figure \ref{fig:BaltimorePort}.

It is worth noting that suggested approach for constructing isochrones via density-based clustering/filtering of trajectory data, yields a different measure of accessibility than isochrones calculated from travel times. The proposed isochrones would encompass locations where many people \textit{have traveled} to within a specified time period, whereas the latter show locations which people \textit{could} reach in the same period of time. Accordingly, some relatively close but less-visited (perhaps unsafe or unpopular) areas may be excluded from the trajectory-based isochrones, thus providing a different picture of accessibility to various facilities (e.g., supermarkets, gas stations). Another advantage of designing isochrones based on trajectory data is that it can be carried out without information about the transportation network and historical travel times along various road links.

While the proposed density-based clustering approach represents an innovative application of trajectory data to quantify mobility, DBSCAN's results are very sensitive to the input parameters, where the best input parameters depend heavily on the size and specific distribution of the dataset (e.g., note different parameter values reported in Figure \ref{DBSCANparameters} and different number of points mentioned in the previous paragraph). Consequently, the proposed approach suffers from excessive parameter tuning and the need for visual sanity checks. In particular, parameter setting includes a trial and error approach with the goal of having DBSCAN provide a large cluster of points around the origin location that neither encompasses remote waypoints nor excludes areas with many waypoints (see Figure \ref{fig:DBSCAN} for an example). As an extension of the proposed approach, one could try to develop a method to automatically adjust parameter setting for different case studies. Development of such a method would certainly represent a challenging task.

\subsection{Public transit}
\begin{figure}
	\begin{subfigure}[t]{0.23\textwidth}
		\frame{\includegraphics[width=1\textwidth]{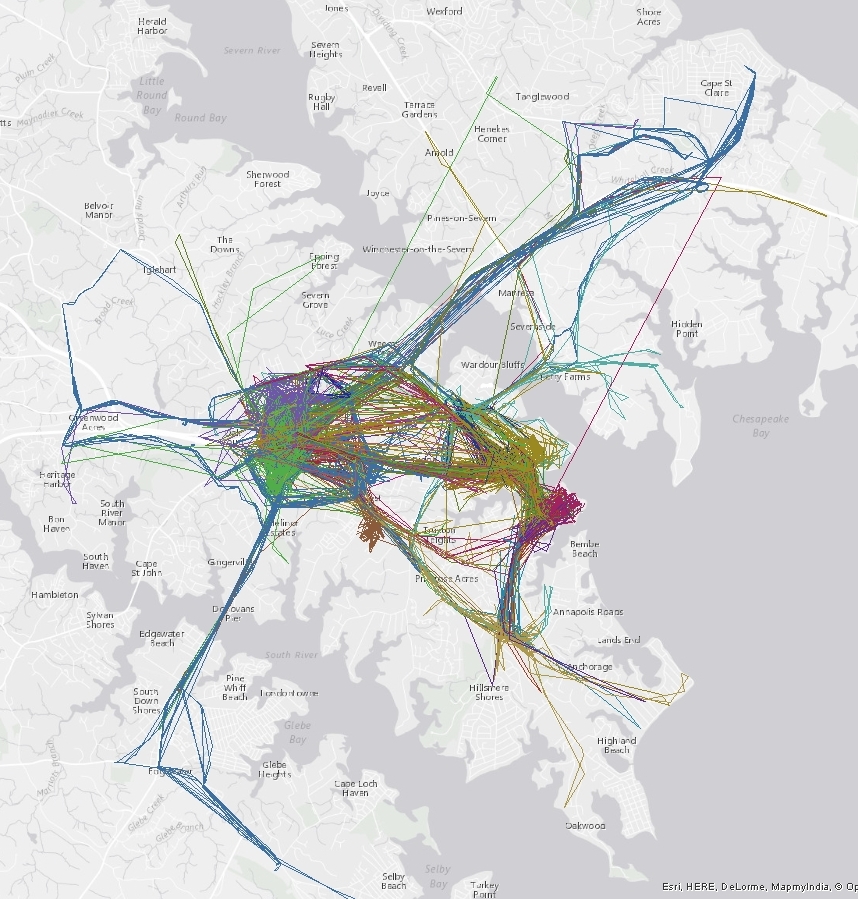}}
		\caption{O-D clusters and their raw trajectories}\label{fig:PublicTransitb}
	\end{subfigure}\hfill
	\begin{subfigure}[t]{0.23\textwidth}
		\centering
		\frame{\includegraphics[width=1\textwidth]{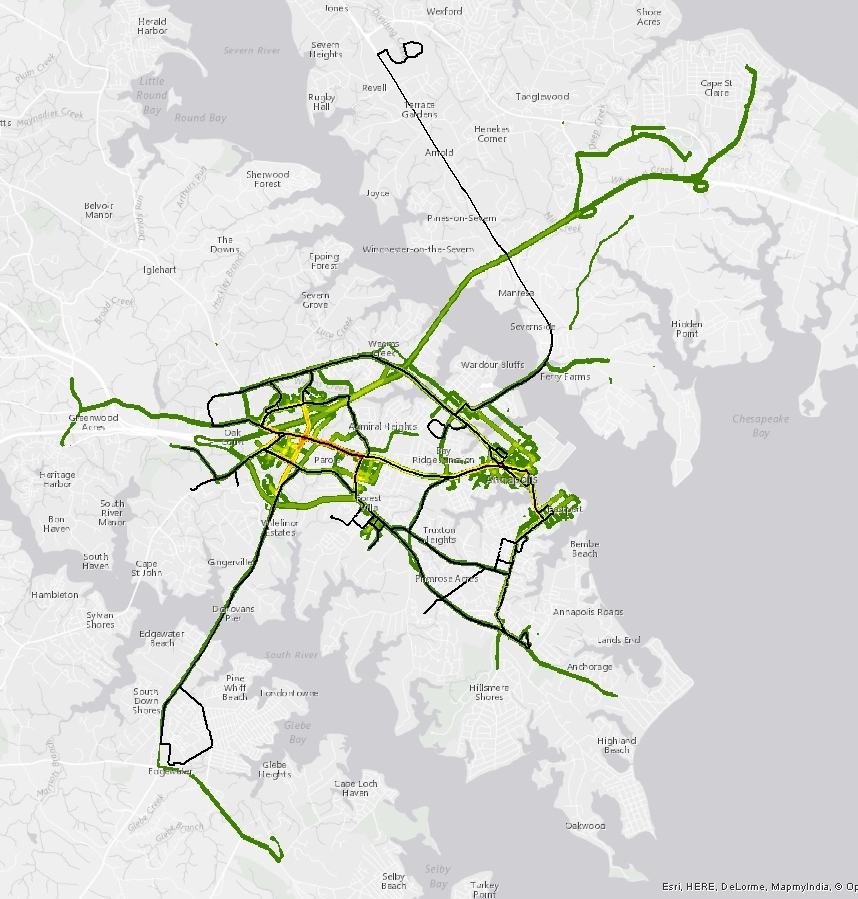}}
		\caption{Heat map of trajectories vs. transit network}\label{fig:PublicTransitc}
	\end{subfigure}%
	\caption{Clusters of trips in Annapolis can be used to modify bus transit network in order to accommodate additional movements. The last visual contrasts commonly traveled routes with the transit system shown with solid black line.}\label{fig:PublicTransit}
\end{figure}

Public transit operates most efficiently when it provides services that appropriately match customers' spatial and temporal demand. Since GPS traces capture spatio-temporal patterns, they can be used to improve public transit by comparing existing transit routes with actual trips in a metropolitan region. To illustrate this application, we focus on trips in the Annapolis, MD region and cluster their O-D pairs using the OPTICS algorithm \cite{Ankerst1999optics}. The clustered O-D pairs are color-coded and shown in Figure \ref{fig:PublicTransitb}. The map-matched trajectories are then overlaid onto the existing Annapolis transit network in Figure \ref{fig:PublicTransitc}, applying a linear heat map in order to emphasize the most-traveled routes. This visual comparison of important trajectories and the transit network reveals that some highly-traveled routes are currently not covered with the transit system. This simple visual comparison may be useful for facilitating discussion with the City of Annapolis about modifying bus routes to best accommodate additional trips. Furthermore, given sufficient interest in a full transit system evaluation or re-design, the GPS traces could be used in conjunction with an array of data mining, operations research and microsimulation techniques to explore spatio-temporal characteristics of trips, optimize routes and service frequencies, and evaluate potential savings.

\subsection{Safety}
\begin{figure}
	\centering
	\begin{subfigure}[t]{.24\textwidth}
		\centering
		\frame{\includegraphics[height=31mm]{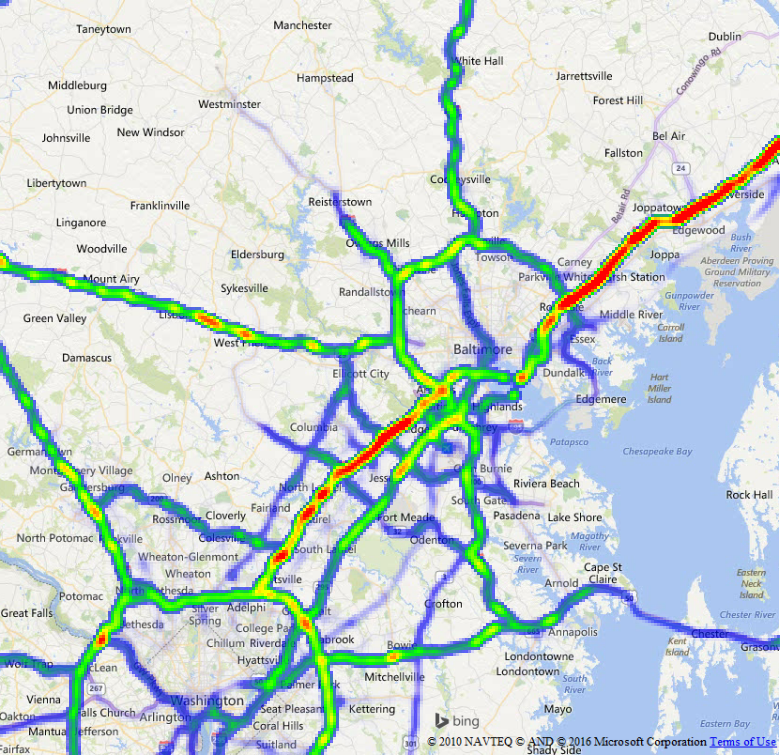}}
		\caption{Washington-Baltimore}\label{WashingtonBaltimoreSpeeds}
	\end{subfigure}\hfill
	\begin{subfigure}[t]{0.24\textwidth}
		\centering
		\frame{\includegraphics[height=31mm]{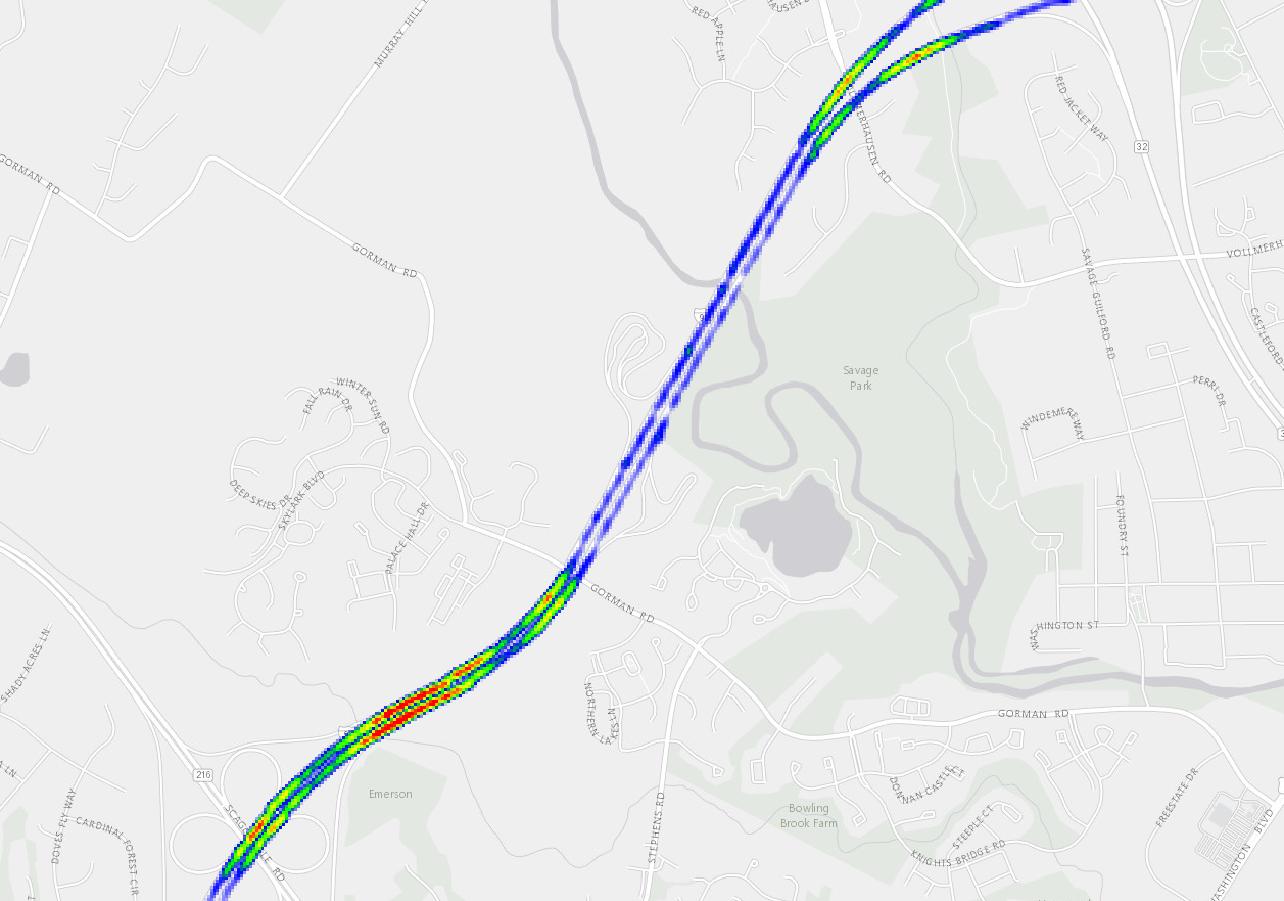}}
		\caption{I-95 nearby North Laurel}\label{I95Speeds}
	\end{subfigure}
	\caption{Heat map of locations with higher speed recordings indicates candidate locations for implementation of speed cameras. After an initial analysis at the regional level (Figure \ref{WashingtonBaltimoreSpeeds}), an analyst can focus on a particular road segment and explore directional speed profiles (Figure \ref{I95Speeds}). Color thresholds can be changed to narrow down candidate locations.}\label{SpeedViolations}
\end{figure}

Detailed trajectory data can reveal speed profiles of millions of \textit{anonymized} drivers, which has important safety implications. We compute average speeds between all consecutive waypoints in our data set (which includes 1.4 billion GPS points), and focus on ones with higher than average speeds. Figure \ref{SpeedViolations} shows a heat map that indicates locations where higher speeds are recorded with greater frequency. The result could be readily used by agencies in charge of deploying speed cameras and radar patrols, which would likely help improve safety and reduce property damage. However, we stress here that trajectory data is anonymized and speeding cannot be traced back to individuals; the goal is to identify segments of the road network that may be good candidates for safety improvements.

\subsection{Weight control}
Some truckers may overload their vehicles in order to increase their productivity and profits, which results in excessive pavement and environmental damages. An effective way of reducing this damage is to implement weigh-in-motion (WIM) systems, which are designed to detect and fine overweight trucks. However, an issue with these systems is that they are inroad facilities, which once deployed in a transportation network remain in their locations for several years. Thus, truckers quickly learn the locations of these systems and can start taking detours in order to avoid them, which can lead to increased pavement and environmental damage due to more vehicle miles traveled \cite{markovic2015evasive,markovic2017evasive}. 

\begin{figure}
	\begin{subfigure}{.24\textwidth}
		\frame{\includegraphics[height=29mm]{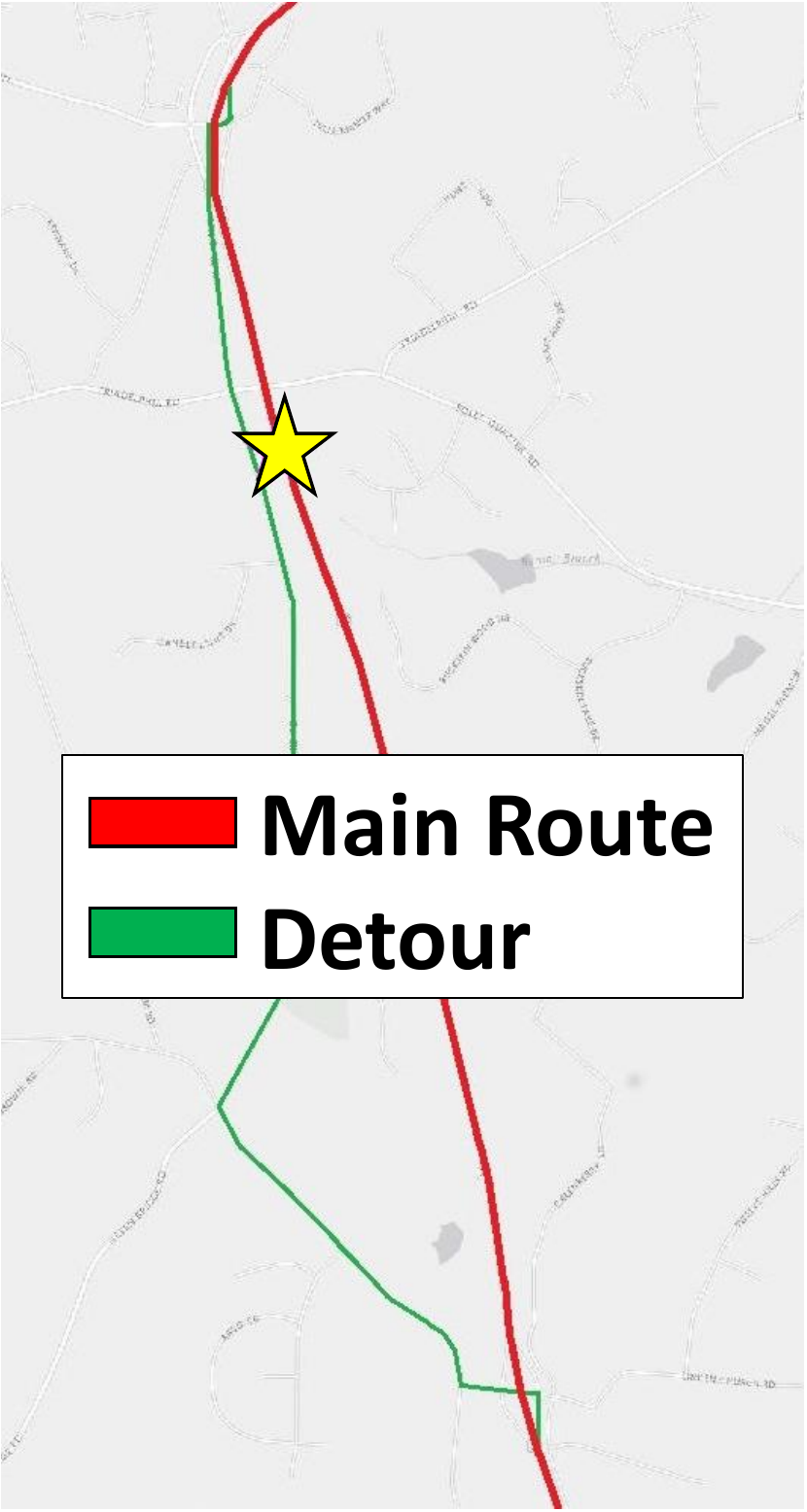}} \,	
		\frame{\includegraphics[height=29mm]{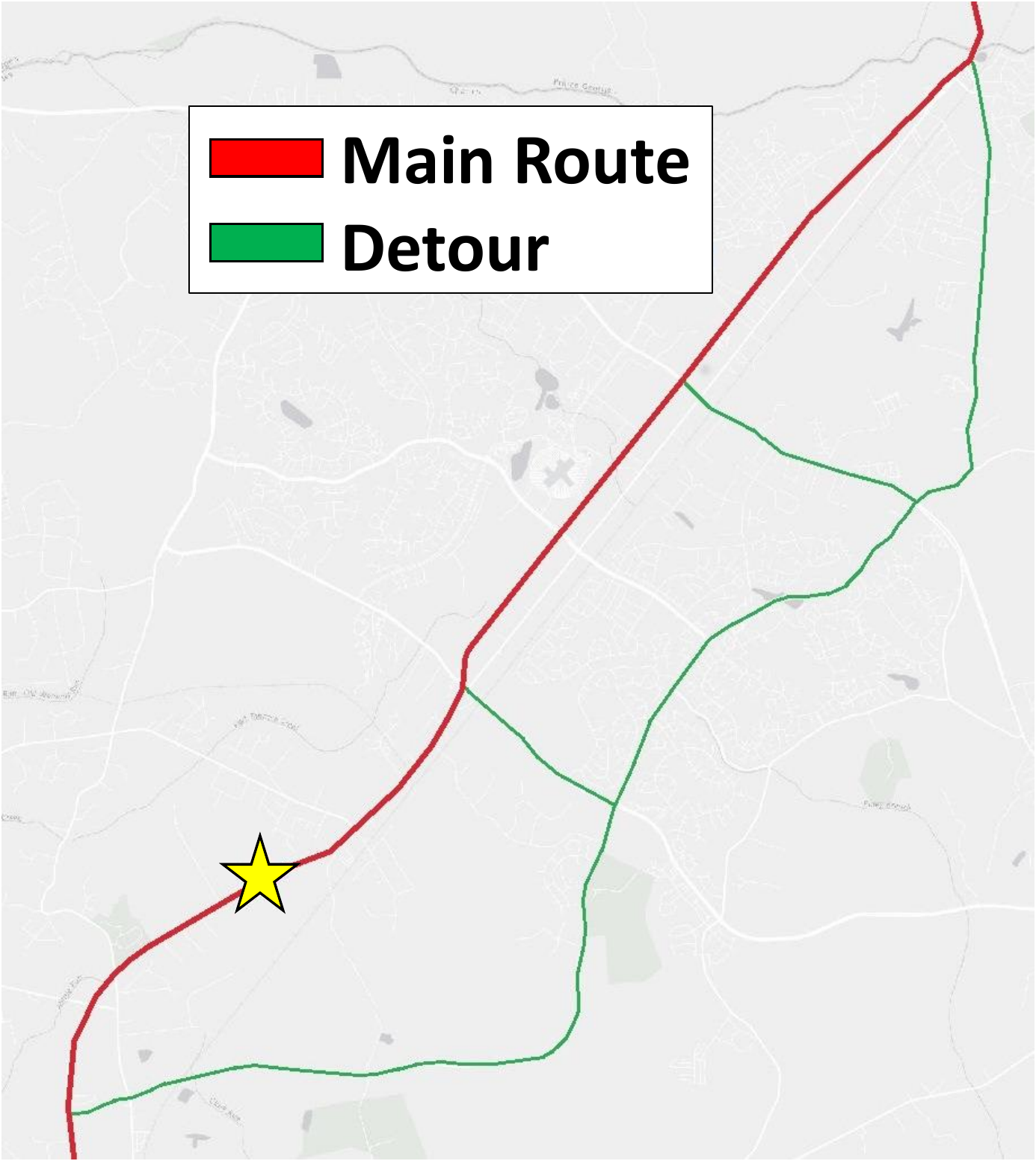}}
	\end{subfigure}\hfill
	\begin{subfigure}{.24\textwidth}
	\scalebox{0.9}{
		{\scriptsize
			\bgroup
			\def\arraystretch{1.2}% 
			\begin{tabular}{ccc}\hline
				Weight Class & Main Road & Circumvent \\
				(1,000 lb) & (veh) & (\%) \\ \hline \hline
				$[0, 14]$ & 3,794 & 1.45 \\
				$(14, 26)$ & 12,333 & 0.61 \\
				$[26, \infty)$ & 4,847  & 0.00	\\ \hline\hline
				$[0, 14]$ & 1,775 & 1.82 \\
				$(14, 26)$ & 6,717 & 1.03 \\
				$[26, \infty)$ & 6,403  & 0.02 \\\hline
				
			\end{tabular}	
			\egroup
		}
	}
	\end{subfigure}
	\caption{Examining potential evasion of WIM systems at MD 32 East (left) and US-301 North (right) along immediate detours. Locations of WIM systems along the main routes are indicated with pentagrams.}\label{WIMevasion}
\end{figure}

Trajectory data can reveal route choices of millions of \textit{anonymized} drivers, which can be used to investigate the extent to which truckers are avoiding WIM systems. As an illustrative case study, we consider two WIM systems in Maryland and compute the percentage of vehicles that take immediate detours (Figure \ref{WIMevasion}). The results indicate that trucks above 26k LB are not bypassing the systems, whereas $0.6\%-1.8\%$ of other vehicles are deviating from the main road in the immediate vicinity of the WIM systems and then returning to the main road afterwards. This may suggest an evasion problem, because at least one third of these trips incurred greater travel times by taking detours. Also, in our data we are unable to differentiate between passenger cars that would not have an incentive to avoid WIM systems and trucks below 14k lb, so the percent of small trucks taking detours may be much higher. It is noteworthy that considering additional alternative routes would provide a better picture of potential evasive strategies. Again, we stress that trajectory data is anonymized and potential evasions cannot be traced back to individuals; the objective is to identify areas that may be good candidates for additional weight control, which would reduce excessive damages and also improve safety for all the road users.

\section{Discussion}
Since characteristics of trajectory data can significantly influence its applicability and thereby value, we provide a discussion about some possible challenges that transportation agencies should be aware of when purchasing trajectory data. The following is a list of potential data-related issues that agencies may want to discuss with data vendors in order to obtain a more complete picture about applicability of a specific dataset. Some general recommendations to transportation agencies interested in acquiring trajectory data are included as well.

\subsubsection{Sampling rate} 
The average time lapse between consecutive waypoints significantly affects applicability of trajectory data, and agencies should try to acquire data with the highest granularity possible (e.g., with the median or average time lapse of 1 second). For example, a large time lapse between waypoints may not influence estimation of O-D matrices, but it could make reconstruction of road-based trajectories a significant challenge, especially in dense urban areas where it may be impossible to determine which route a vehicle took. Thus, it is important to request information about the granularity of data and assess how it would influence the anticipated analysis before actually acquiring data. Also, requesting road-based trajectories in addition to raw data, may save agencies quite a bit of time and resources needed for map matching, which was discussed in Section \ref{sec:mapmatching}.
	
\subsubsection{Spatial precision} 
The number of decimal numbers used to report waypoint latitudes/longitudes is another factor that can influence applicability of trajectory data. For example, rounding a waypoint location to four decimal numbers introduces an error of about 11 m. While this error would not necessarily prevent us from reconstructing road-based trajectories or studying demand, it would significantly affect speed estimates and its use in microsimulation models. Assuming that the median spacing between two consecutive waypoints is 28 m (Figure \ref{WaypointsStat}), location errors of 11 m would make speed estimates meaningless. The same applies to computing vehicle acceleration/deceleration rates that are needed for microsimulation models used to estimate emissions, such as VT-Micro \cite{Rakha2004}. Thus, agencies should request latitudes/longitudes expressed with six decimal numbers, and still account for the errors that are inherent to GPS technology. 
	
\subsubsection{Division of trajectories into trips} 
Transportation agencies should be aware that GPS companies may reset a trip whenever the vehicle is idle for a specified period of time (e.g., 10 minutes). When this occurs within the boundaries of a state for which data was purchased, an analyst can still chain consecutive trips by looking at the unique device identifications. However, when a trip gets reset once it leaves the state, than the information about subsequent lags of the trip is lost. This is probably the reason that Figure \ref{WaypointsStat} does not include any trips going to the West Coast, as such a long trip would necessitate stops long enough to reset the trip. To overcome this problem and gain better insight into long-distance trips, agencies from multiple states could jointly purchase data for an entire region (e.g., East Coast or all of USA), which also may be more cost efficient due to economies of scale.
	
\subsubsection{Population bias} 
Transportation agencies should be aware of the bias in data towards certain types of vehicles. For example, the dataset discussed in this paper is biased towards delivery trucks (Figure \ref{TripAttributes}). This may not represent a major issue if the observed region includes a network of ATR stations that can differentiate between different vehicle types. In this case, an analyst can determine the penetration rates of different types of vehicles (passenger cars vs. trucks) and account for any bias in further analysis. However, when such a network of sensors is unavailable, correcting for the bias becomes a challenge and may limit applications of trajectory data (e.g., estimation of an O-D matrix becomes a challenge). Therefore the government agencies interested in purchasing trajectory data should also account for the availability of other data sources that would enable them to correct for the aforementioned bias in data.
	
\subsubsection{Unique device identifications} 
Each trip in a trajectory dataset includes an identification (ID) of the device it was recorded from. Device IDs enable an analyst to chain consecutive trips of the same vehicle and thereby reconstruct its movement over a longer period of time, which provides a better insight into mobility patterns. However, data vendors may decide to periodically change device IDs (e.g., at midnight) for privacy or some other reasons, which clearly limits the analysis. Thus, transportation agencies interested in purchasing trajectory data should inquire about vendor's policies with respect to resetting device IDs and account for its implications on their analyses. Additional issues that analysts should be aware of are occasionally duplicated or swapped device IDs, which may arise when resetting device IDs. These and other issues related to trajectory data are discussed in \cite{andrienko2016understanding}.

\section{Conclusions}
This paper synthesizes innovative applications of trajectory data in road transportation, which is relevant to government agencies looking to introduce this type of data into their analyses and decision making processes. We provide a literature review illustrating applications of trajectory data in six areas of road transportation systems analysis: demand estimation, modeling human behavior, designing public transit, traffic performance measurement and prediction, environment and safety. Additionally, we perform an extensive analysis of 20 million GPS trajectories in Maryland, demonstrating both existing and new applications of trajectory data in transportation. We employ an array of techniques encompassing data processing and management, machine learning, and visualization, and describe best-practices for using them to extract value from trajectory data, thus allowing transportation agencies to estimate the time and effort needed to introduce this type of data into their modeling efforts. As trajectory data becomes more prevalent and acquisition costs decrease, we believe that this type of data will become an invaluable resource to transportation agencies across the world. 

\section*{Acknowledgment}
The authors would like to thank Subrat Mahapatra and the Maryland State Highway Administration for their support throughout this project. Help from the I-95 Corridor Coalition and the City of Annapolis are also appreciated. The last two authors also acknowledge support by EU projects VaVeL ``Variety, Veracity, VaLue: Handling the Multiplicity of Urban Sensors''
(grant agreement 688380) and Track\&Know ``Big Data for Mobility Tracking
Knowledge Extraction in Urban Areas'' (grant agreement 780754). This support is gratefully acknowledged, but it implies no endorsement of the findings.

\ifCLASSOPTIONcaptionsoff
  \newpage
\fi



%\bibliographystyle{IEEEtran}

\bibliography{od_ref}

%\begin{biography}[{\includegraphics[width=1in,height=1.25in,clip,keepaspectratio]{mshell}}]{Michael Shell}

\begin{IEEEbiography}[{\includegraphics[width=1in,height=1.25in,clip,keepaspectratio]{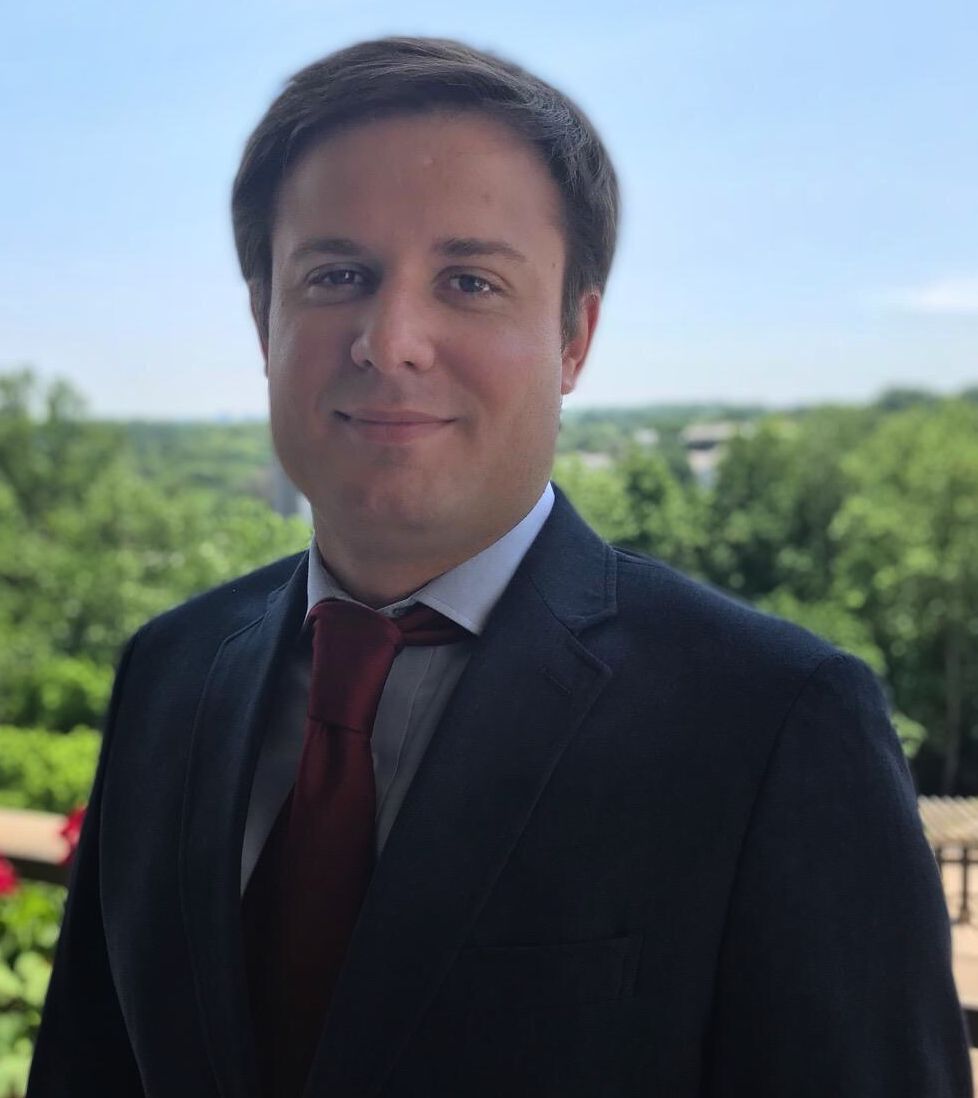}}]{\\Nikola Markovi\'c} received his Ph.D. degree in transportation engineering from the University of Maryland in 2013. His research interests include applications of operations research and machine learning in transportation systems analysis. Currently, he is working at the Center for Advanced Transportation Technology, University of Maryland, USA.
\end{IEEEbiography}
\begin{IEEEbiography}[{\includegraphics[width=1in,height=1.25in,clip,keepaspectratio]{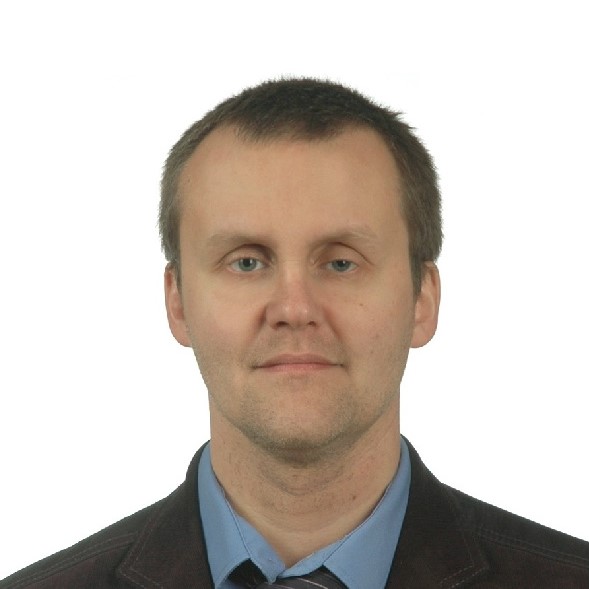}}]{Przemys\l{}aw Seku\l{}a}
received his Ph.D. degree in management from the University of Economics in Katowice in 2012. His research interests include applications of machine learning and artificial intelligence in transportation. Currently, he is working as a researcher at the Center for Advanced Transportation Technology, University of Maryland, USA, and an Assistant Professor at the University of Economics in Katowice, Poland.
\end{IEEEbiography}
\begin{IEEEbiography}[{\includegraphics[width=1in,height=1.25in,clip,keepaspectratio]{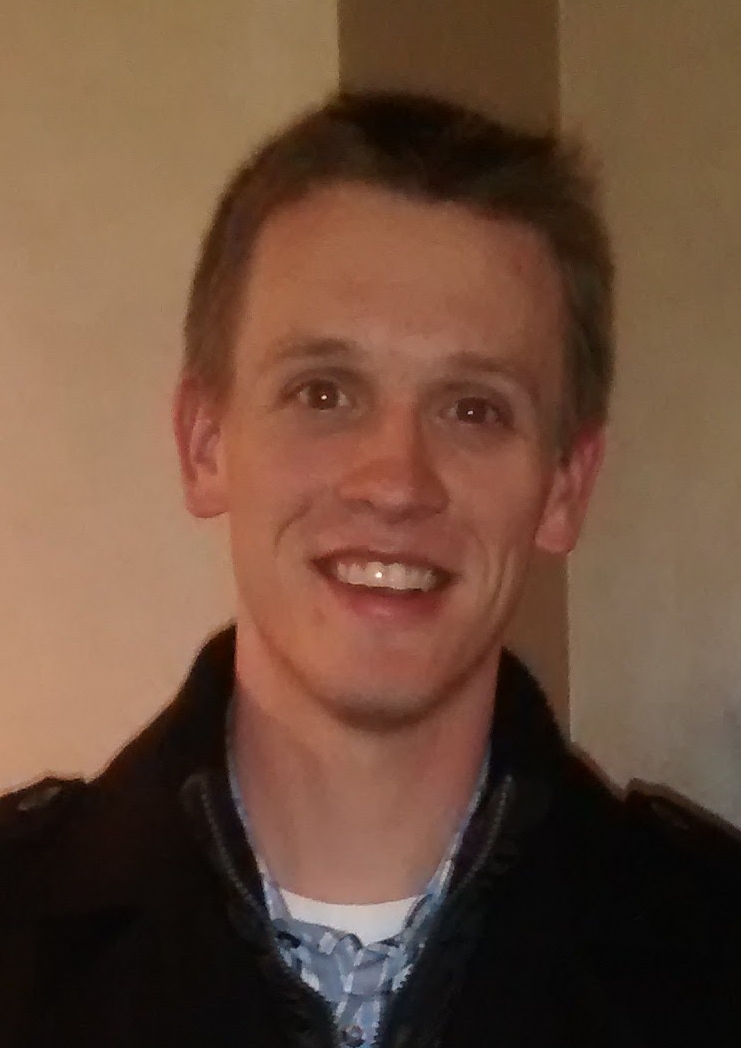}}]{Zachary Vander Laan} received his M.S. degree in civil engineering from the University of Maryland in 2017. His research interests include intelligent transportation systems, data visualization, and applications of machine learning in transportation. Currently, he is working at the Center for Advanced Transportation Technology, University of Maryland, USA.
\end{IEEEbiography}
\begin{IEEEbiography}[{\includegraphics[width=1in,height=1.25in,clip,keepaspectratio]{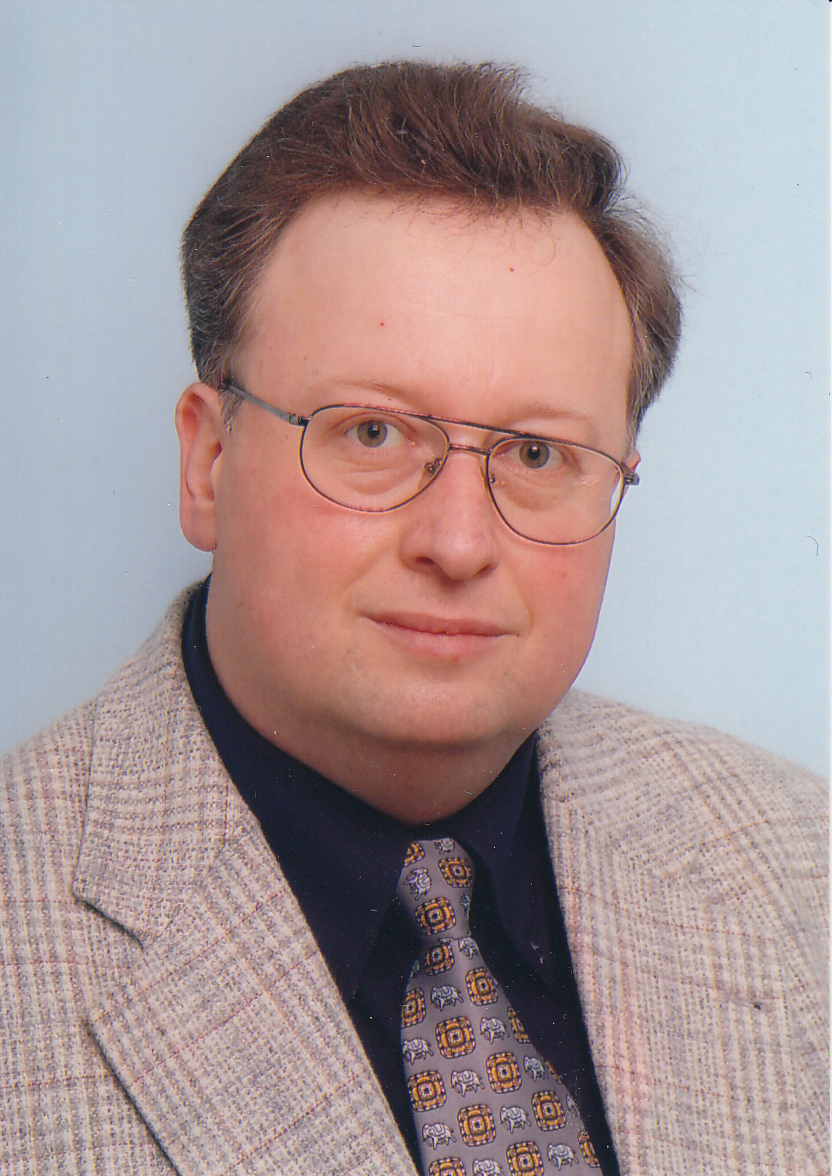}}]{Gennady Andrienko} is a Lead Scientist responsible for the visual analytics research with Fraunhofer Institute Intelligent Analysis and Information Systems and a Professor (part-time) with City University
London. He has co-authored two monographs, Exploratory Analysis of Spatial and Temporal Data (Springer, 2006) and Visual Analytics of Movement (2013), and more than 80 peer-reviewed journal papers. From 2007 to 2015, he was chairing the ICA
Commission on GeoVisualization. He co-organized scientific events on visual analytics, geovisualization, and visual data mining, and co-edited 13 special issues of journals.
\end{IEEEbiography}
\begin{IEEEbiography}[{\includegraphics[width=1in,height=1.25in,clip,keepaspectratio]{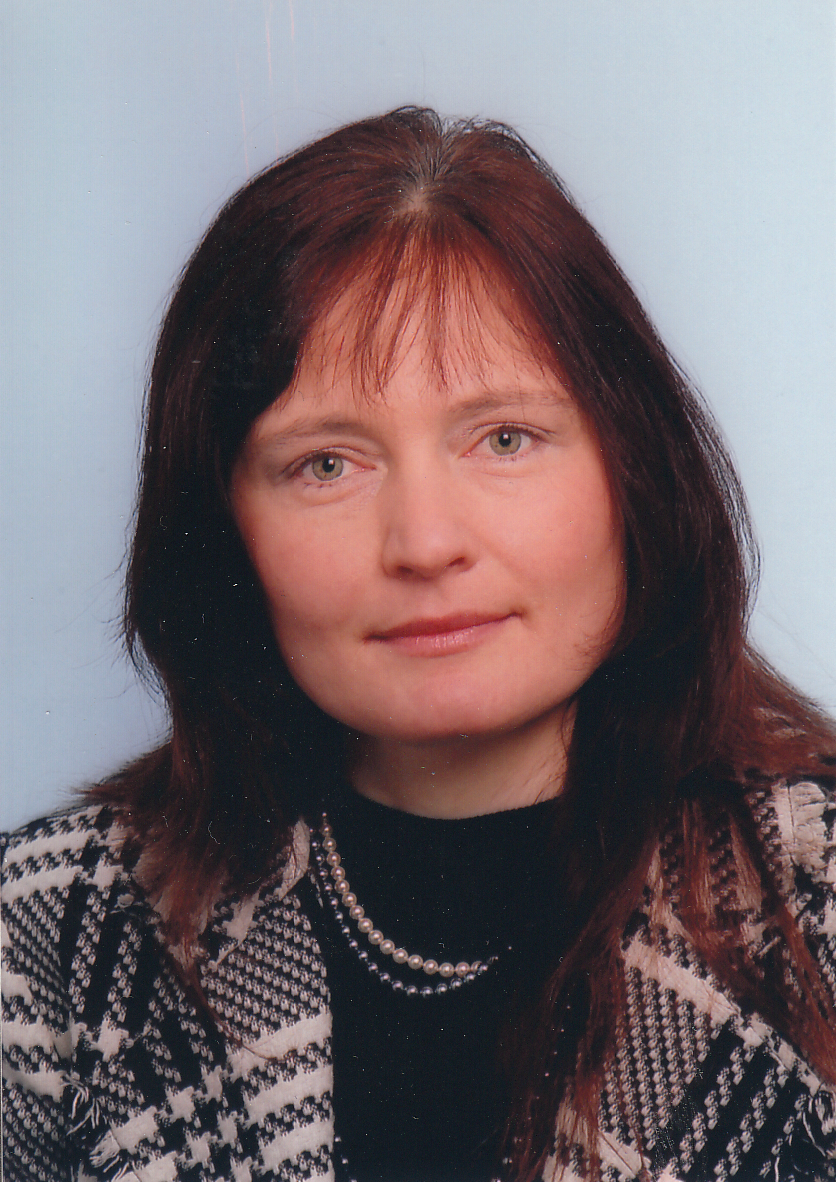}}]{Natalia Andrienko} has been with GMD, currently 	Fraunhofer Institute Intelligent Analysis and Information Systems, since 1997. Since 2007, she has been a Lead Scientist, where she has been involved in visual analytics research. Since 2013, she has been a Professor (part-time) with City University London. She has co-authored the monographs Exploratory Analysis of Spatial and Temporal Data (Springer, 2006) and Visual Analytics of Movement	(Springer, 2013) and over 70 peer-reviewed journal papers. She received best paper awards at AGILE 2006, EuroVis 2015, and IEEE VAST 2011 and 2012 conferences; best poster awards at AGILE 2007, ACM GIS 2011, and IEEE VAST 2016; and VAST challenge awards 2008 and 2014.
\end{IEEEbiography}


%\vfill


%\vspace*{\fill}

\end{document}